\documentclass[
]{ceurart}

\usepackage{listings}
\usepackage{placeins}
\usepackage{graphicx}
\usepackage{array}
\usepackage{booktabs}
\usepackage{multirow}
\usepackage{adjustbox}

\ExplSyntaxOn
\cs_set:Npn \__reset_fig: {
  \tl_set:Nx \l_fig_pos_tl { t }
  \tl_set:Nx \l_fig_cols_tl { 1 }
  \tl_set:Nn \l_fig_align_tl { \raggedright }
  \skip_set:Nn \l_fig_abovecap_skip { 6pt }
  \skip_set:Nn \l_fig_belowcap_skip { 6pt }
  \skip_set:Nn \l_fig_abovefig_skip { 6pt }
  \skip_set:Nn \l_fig_belowfig_skip { 6pt }
}
\ExplSyntaxOff

\ExplSyntaxOn
\RenewDocumentCommand \printorcid { } {
  \group_begin:
  \int_compare:nTF { \g_orcid_int > 0 } {
    \tex_let:D \thefootnote \relax \footnotetext {
      \raggedright
      \hspace*{-\parindent}
      \hspace*{-\footnotemargin}
      \bool_if:NTF \g_ceur_nologo_bool
      { \textsc{orcid:\c_space_token} }
      { \faIcon{orcid}\c_space_token }
      \seq_use:Nn \g_ceur_orcid_seq { ;~ }
    }
  } { }
  \group_end:
}
\ExplSyntaxOff

\newcommand{\scorecamimg}[2]{%
  \IfFileExists{#2}{\includegraphics[width=#1,height=#1,keepaspectratio]{#2}}{\fbox{\parbox[c][#1][c]{#1}{\centering\scriptsize missing\\{\ttfamily\detokenize{#2}}}}}%
}
\begin{document}

\copyrightyear{2026}
\copyrightclause{Copyright for this paper by its authors.
  Use permitted under Creative Commons License Attribution 4.0
  International (CC BY 4.0).}

\conference{CLEF 2026 Working Notes, September 21-24, Jena, Germany}

\title{Calibrated Similarity and Graph Clustering for Open-Set Animal Re-Identification}

\author[1, 3]{Mohamed ElBassat}[%
email=cds.MohamedMohamed24510@alexu.edu.eg,
orcid=0009-0006-3917-1319
]
\cormark[1]
\fnmark[1]

\author[1,2]{Seifeldin Elkerdany}[
email=seifeldean.hosny.2024@aiu.edu.eg,
orcid=0009-0005-7199-7391
]
\cormark[1]
\fnmark[1]

\author[1,5]{Mohamed ElBialy}[
    orcid = 0009-0001-2281-7359
]
\author[1,2]{Gamal Abouelhamd}[
    orcid = 0009-0004-7223-5260
]
\author[1,4]{Jana Ghoneim}[
    orcid = 0009-0008-9943-295X
]
\author[1,3]{Assem Elkady}[
    orcid = 0009-0003-9847-9944
]
\author[1,3]{Mohamed Elboraay}[
    orcid = 0009-0002-1018-8150
]
\author[6]{Nelly Semenova}[
    orcid = 0000-0002-0190-8382
]
\fnmark[2]

\address[1]{Made In Alexandria Artificial Intelligence Team, Alexandria, Egypt}
\address[2]{Faculty of Computer Science and Engineering, Alamein International University, New Alamein City, 51718, Egypt}
\address[3]{Faculty of Computers and Data Science, Alexandria University, Alexandria, Egypt}
\address[4]{Faculty of Engineering, Alexandria University, Alexandria, Egypt}
\address[5]{Alexandria Higher Institute of Engineering and Technology, Alexandria, Egypt}
\address[6]{Moscow Pedagogical State University (MPGU University), 1/1 Malaya Pirogovskaya St., Moscow, 119435, Russian Federation}

\cortext[1]{Corresponding authors.}
\fntext[1]{These authors contributed equally.}
\fntext[2]{This author contributed by providing access to the previous year's second-place solution and computational resources; they did not contribute to the development of the proposed method.}

\begin{abstract}
AnimalCLEF26 addresses discovery-oriented animal re-identification, where systems must both attach query images to known individuals and discover unseen individuals by clustering them correctly. We present a similarity-to-clustering pipeline for this setting across Eurasian lynx, fire salamander, loggerhead sea turtle, and Texas horned lizard images. The method first isolates the target specimen using segmentation and then applies lightweight species-specific preprocessing for lynx, sea turtle, and salamander images to enhance identity-relevant visual cues, while Texas horned lizard images are used after segmentation only. Pairwise similarities are then estimated with WildFusion by calibrating and combining a MiewID global descriptor with two local matching branches, ALIKED + LightGlue and DISK + LightGlue. The resulting query--query similarities are refined and converted into identity clusters using graph-based clustering, while query--database similarities are used to attach confident samples to known identities. We evaluate training-free and fine-tuned MiewID variants, including Dynamic ArcFace and SphereFace2-Focal adaptations, and combine them in the final ensemble. Our selected ensemble substantially improves on the WildFusion baseline, achieving our best public ARI of 0.72124 and a private ARI of 0.70393, while a simpler preprocessing-before-calibration variant achieves our best private ARI of 0.71087. These results indicate that calibrated global-local fusion with species-aware preprocessing choices is effective for open-set wildlife re-identification under challenging field conditions and visual variation. The implementation code is available on \href{https://github.com/MIA-AI-Team/AnimalCLEF26?tab=readme-ov-file}{GitHub}.
\end{abstract}


\begingroup
\hfuzz=500pt
\maketitle
\endgroup

\begin{figure}[!htbp]
    \centering
    \includegraphics[width=0.95\linewidth]{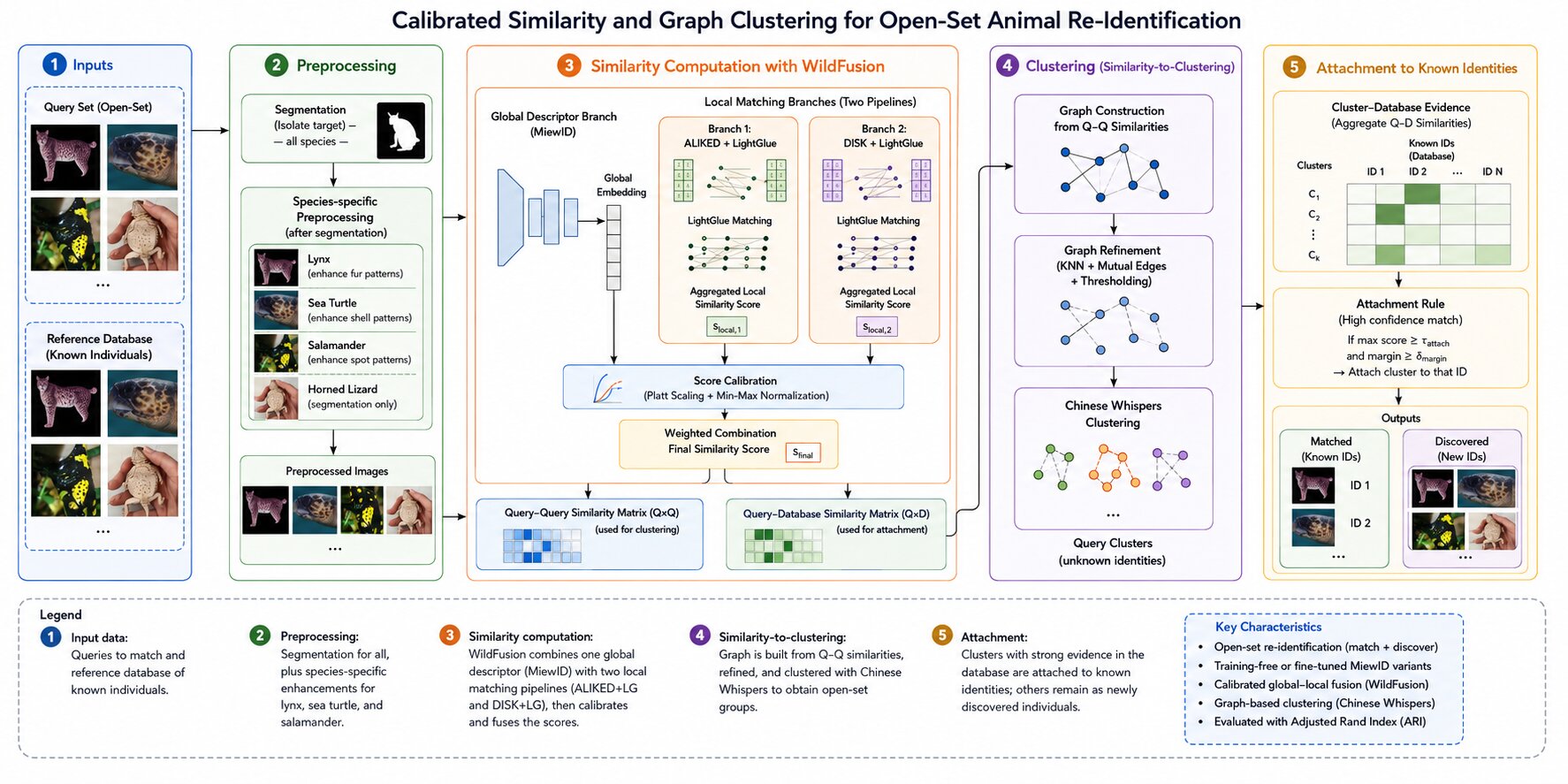}
    \caption{Overview of the proposed similarity-to-clustering pipeline.}
    \label{fig:pipeline_overview}
\end{figure}

\section{Introduction}

Animal re-identification supports wildlife monitoring by linking repeated observations to the same individual for tasks such as population estimation, movement tracking, and behavioral analysis \cite{vcermak2024wildlifedatasets, adam2024wildlifereid10k}. Manual identification and physical or electronic marking remain useful, but they are labor-intensive and scale poorly to the growing volume of images collected from camera traps, drones, and online sources \cite{zhang2024animalsurvey, machado2021photoid, adam2024wildlifereid10k}. This motivates the development of automated animal re-identification systems \cite{vcermak2024wildlifedatasets, adam2024wildlifereid10k}. In this work, we address this problem in the context of AnimalCLEF 2026, which was organized as part of the LifeCLEF 2026 lab and focuses on discovery and re-identification of individual animals in an open-set setting \cite{lifeclef2026, animalclef2026overview}.

In practice, this problem is not purely closed-set. A useful system must both match images of known individuals and discover previously unseen individuals by grouping their observations correctly \cite{animalclef2026, adam2024wildlifereid10k}. This open-set setting is challenging because the same animal may appear under different poses, viewpoints, lighting conditions, occlusions, ages, and backgrounds, while different individuals of the same species may differ only in subtle visual patterns \cite{animalclef2026,adam2024wildlifereid10k}.

AnimalCLEF26 provides a benchmark for this setting across Eurasian lynx, fire salamander, loggerhead sea turtle, and Texas horned lizard images, and evaluates predictions with the Adjusted Rand Index (ARI) \cite{animalclef2026, picek2026czechlynx, biffi2025identification, adam2024seaturtleid2022, hubert1985comparing}. We address the task with a similarity-to-clustering pipeline in two variants: a training-free version and a version that adds task-adapted MiewID representations and ensembling. After segmentation, lynx, sea turtle, and salamander images receive additional species-specific preprocessing, whereas Texas horned lizard images are used after segmentation only. The resulting images are then compared with WildFusion, which combines one global descriptor with two local matching pipelines. Chinese Whispers then clusters the query images, and clusters with strong reference evidence are attached to known identities \cite{cermak2024wildfusion, biemann2006chinese}.

\section{Previous Work}

Recent animal re-identification systems increasingly combine global embedding models, local visual matching, and open-set decision rules. WildFusion introduced a calibrated similarity-fusion framework that combines deep descriptor similarity with local keypoint matching scores, showing that global and local cues are complementary for individual animal identification \cite{cermak2024wildfusion}. This direction was further developed in AnimalCLEF 2025, where top-performing systems adapted global-local fusion to multi-species open-set recognition.

Pakhomov et al. proposed a hybrid global-local pipeline that first used global embeddings for candidate selection, then applied multiple local feature matchers, segmentation-aware filtering, weighted score aggregation, and novelty thresholding. Their system replaced the baseline MegaDescriptor branch with MiewID, added several local matcher combinations, and used segmentation to reduce background noise, leading to the first-place AnimalCLEF 2025 solution \cite{Pakhomov2025IndividualWR}. Semenova proposed a complementary meta-algorithm that combined WildFusion scores with an XGBoost classifier over MegaDescriptor and MiewID features, followed by species-specific Dual-Backbone ArcFace models. This showed that calibrated matching, tabular neighbor context, and metric fine-tuning can be stacked effectively for open-set animal re-identification \cite{semenova2025meta}.

Our method follows the same global-local philosophy, but adapts it to the AnimalCLEF26 discovery setting by combining segmentation, lightweight species-specific preprocessing, WildFusion-based calibrated similarities, fine-tuned MiewID descriptors, and graph-based clustering for open-set identity discovery.

\section{Methodology}

\subsection{Pipeline Overview}

Our method was a similarity-to-clustering pipeline designed for discovery-oriented animal re-identification. Given a set of query images for each species, the pipeline first isolated the target animal through specimen segmentation. It then applied additional species-specific preprocessing for lynx, sea turtle, and salamander images to enhance identity-relevant visual patterns, while Texas horned lizard images were used after segmentation only. Pairwise image similarities were computed using WildFusion, which combined one global descriptor branch with two local matching branches. The local branches were fixed across all official experiments and used ALIKED + LightGlue and DISK + LightGlue, while the global branch was instantiated with MiewID-based descriptors. The complete pipeline is summarized in Figure~\ref{fig:pipeline_overview}.

In the official pipeline, WildFusion calibration mappings were fitted after specimen segmentation and before any additional species-specific preprocessing. After the calibration mappings were fixed, final WildFusion inference was performed on the species-specific preprocessed images for lynx, sea turtle, and salamander, and on the segmentation-only images for Texas horned lizard. This produced two types of similarity matrices: a query--query matrix used for open-set identity discovery through graph-based clustering, and, when labeled database images were available, a query--database matrix used for known-identity attachment.

To improve robustness, we ran the official pipeline with three global descriptor variants: the original pretrained \texttt{MiewID-msv3}, a Dynamic ArcFace fine-tuned MiewID backbone, and a SphereFace2-Focal fine-tuned MiewID backbone. In all three runs, the local matching branches, WildFusion calibration and fusion stage, clustering algorithm, and known-identity attachment logic remained unchanged. The final system combined the outputs of these three descriptor-level pipeline variants.

\subsection{Image Segmentation and Data Preprocessing}

The first stage of the pipeline isolated the target specimen to reduce background noise before feature extraction. We used Segment Anything Model 3 (SAM 3) to generate foreground masks for the target animal, while the lynx images were used directly because the provided crops were already specimen-focused \cite{carion2025sam}. This design follows prior animal re-identification work showing that segmentation and part localization help models focus on individual-specific traits rather than environmental artifacts \cite{adam2024seaturtleid2022,nepovinnykh2022sealid}.

After segmentation, we applied lightweight species-specific preprocessing only when it was expected to improve identity-relevant visual cues \cite{gonzalez2018digital,szeliski2022computer}. For lynx, we enhanced fur texture, markings, and edges using sharpening, HSV-based CLAHE, and bilateral filtering. For sea turtles, we corrected underwater color degradation by boosting the red channel and applying gamma correction to improve shell, head, and scale visibility. For salamanders, we strengthened yellow--black pattern boundaries using Sobel edge enhancement followed by blending and gamma correction. We also explicitly controlled the composited background color after segmentation, since dark salamander regions against a dark background could reduce boundary contrast and make both global descriptors and local keypoint matching less reliable; lighter backgrounds preserved specimen contours more clearly. For Texas horned lizards, we applied no additional preprocessing beyond segmentation, preserving the original ventral spot patterns used for identification \cite{biffi2025identification}.

\begin{figure}[!htbp]
    \centering
    \scriptsize
    \setlength{\tabcolsep}{4pt}
    \begin{tabular}{@{}cc@{}}
        \includegraphics[width=0.46\linewidth]{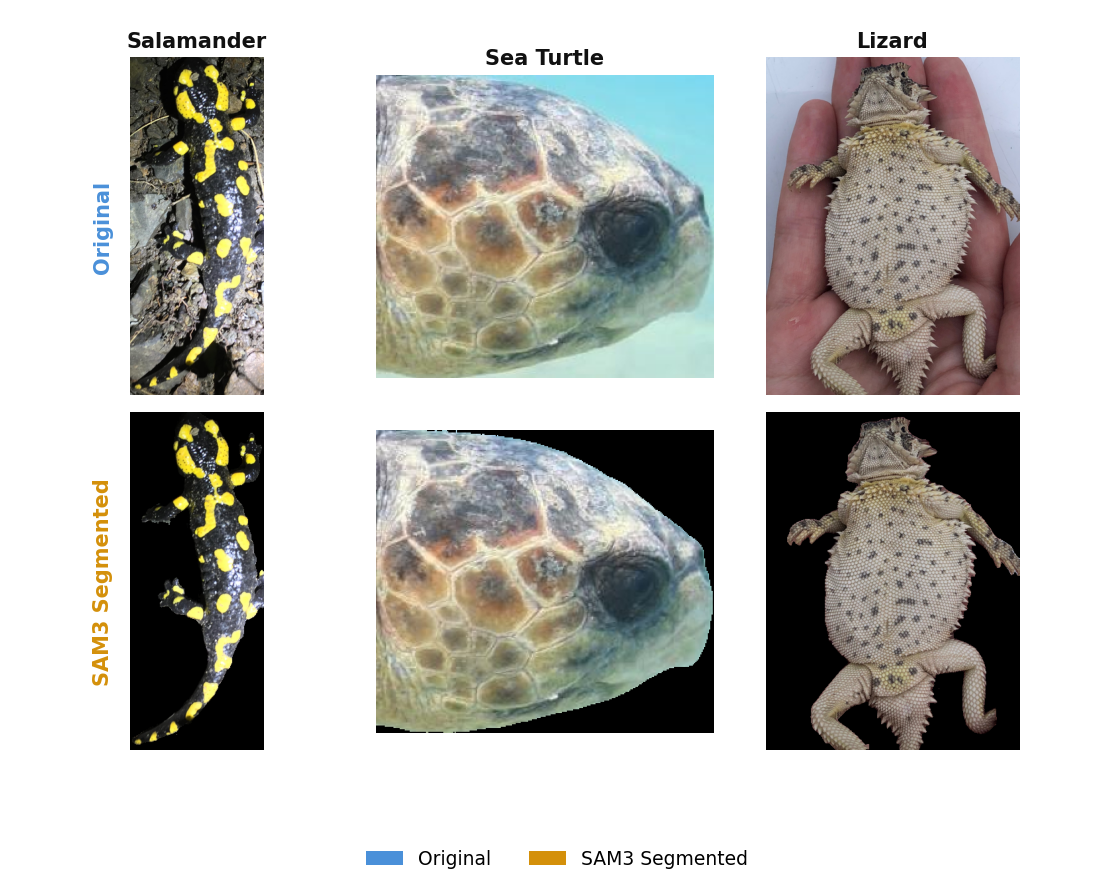} &
        \includegraphics[width=0.46\linewidth,height=0.368\linewidth]{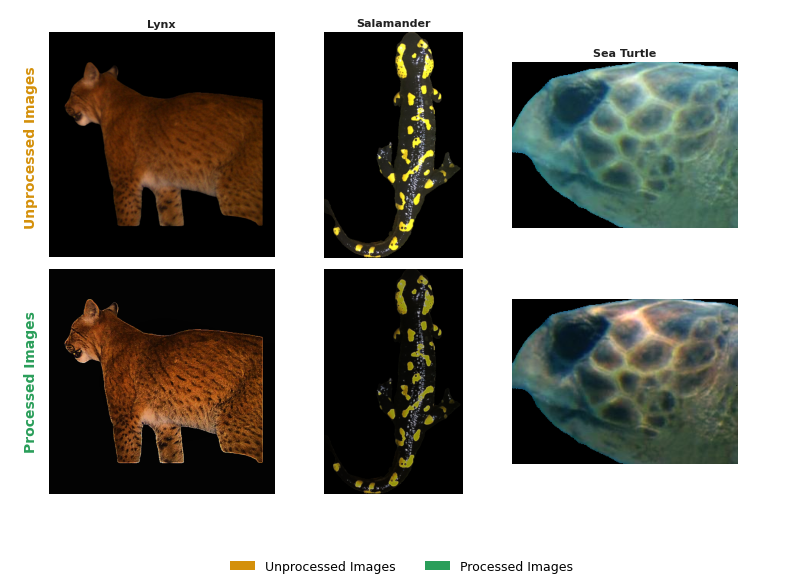} \\
        Specimen segmentation examples & Species-specific preprocessing examples
    \end{tabular}
    \caption{Representative specimen segmentation and species-specific preprocessing outputs.}
    \label{fig:segmentation_preprocessing}
\end{figure}

Representative segmentation and preprocessing outputs are shown in Figure~\ref{fig:segmentation_preprocessing}.

\FloatBarrier

\subsection{Feature Extraction and Similarity Components}

Our pipeline computed image similarity using two complementary sources of visual evidence: a global descriptor branch and two local matching branches. The global descriptor summarized the overall appearance of the segmented animal specimen, while the local matching branches compared fine-grained visual structures such as fur markings, salamander skin patterns, turtle shell texture, scales, spots, and scars. This combination was important in open-set animal re-identification because the same individual could appear under different poses, viewpoints, lighting conditions, and levels of occlusion, while different individuals of the same species could differ only in subtle local patterns.

The local matching branches were fixed across all evaluated pipeline configurations. We used ALIKED + LightGlue and DISK + LightGlue as two complementary local feature pipelines. These branches were used without task-specific fine-tuning and provided local similarity scores that were later calibrated and fused with the global descriptor similarity through WildFusion \cite{cermak2024wildfusion}. In contrast, the global descriptor branch was varied across the main pipeline instances, using either the original pretrained \texttt{MiewID-msv3} descriptor or a fine-tuned MiewID backbone.

\subsubsection{Global Descriptor: MiewID}

The global branch was based on \texttt{MiewID-msv3}, which represents each image with a single embedding and uses pairwise cosine similarity as the global score before WildFusion calibration and fusion \cite{otarashvili2024multispecies}.

MiewID was designed for multi-species animal re-identification, making it suitable for the AnimalCLEF26 setting, where the model had to generalize across visually diverse species and support both known-individual matching and unseen-individual discovery \cite{otarashvili2024multispecies}. In the training-free pipeline, we used the original pretrained \texttt{MiewID-msv3} model directly. In the ensemble setting, the same pipeline was also evaluated with two task-adapted MiewID variants: one fine-tuned with Dynamic ArcFace and one fine-tuned with SphereFace2-Focal loss. In all cases, the global descriptor was the only component changed; the local matching branches, WildFusion calibration and fusion, clustering, and known-identity attachment stages remained unchanged.

\begin{figure*}[t]
\centering
\scriptsize
\setlength{\tabcolsep}{1pt}
\renewcommand{\arraystretch}{0.82}

\resizebox{0.78\linewidth}{!}{%
\begin{tabular}{@{}>{\centering\arraybackslash}m{1.25cm}ccccc@{}}

{\tiny\textbf{Species}} &
{\tiny\textbf{Original}} &
\multicolumn{2}{c}{{\tiny\textbf{MiewID}}} &
\multicolumn{2}{c@{}}{{\tiny\textbf{MegaDescriptor}}}\\

\midrule

{\tiny Lynx} &
\scorecamimg{1.25cm}{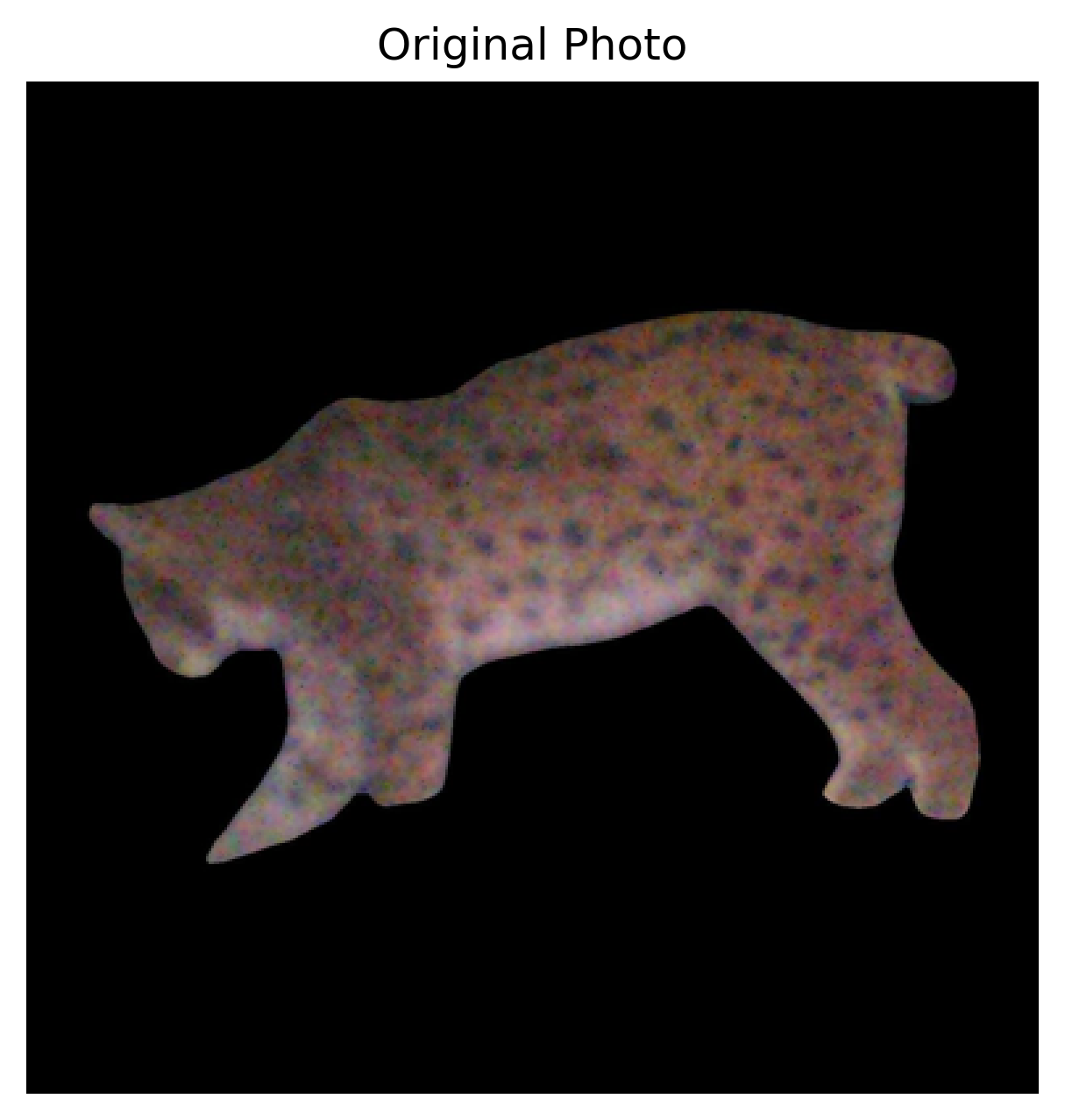} &
\scorecamimg{1.25cm}{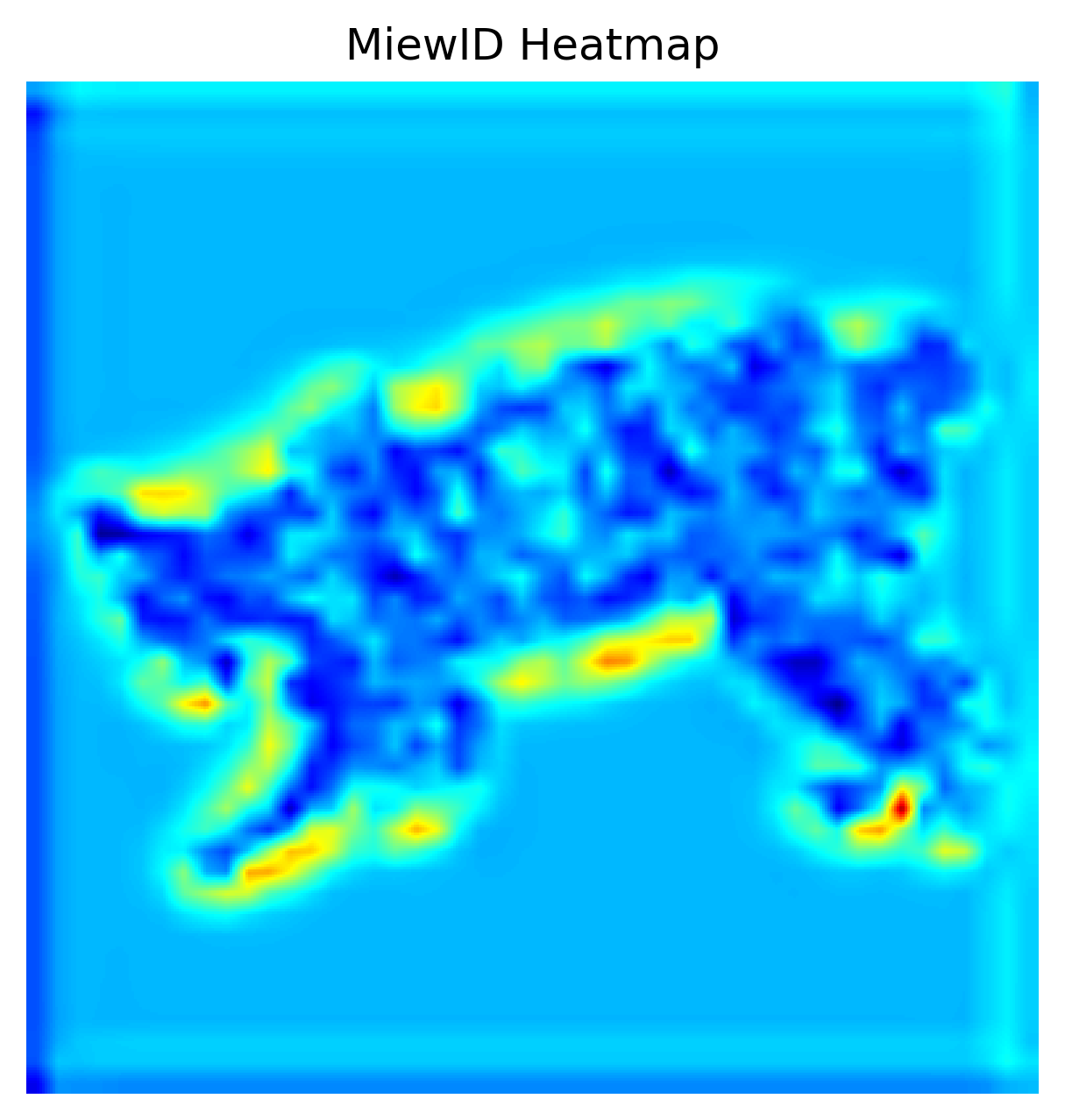} &
\scorecamimg{1.25cm}{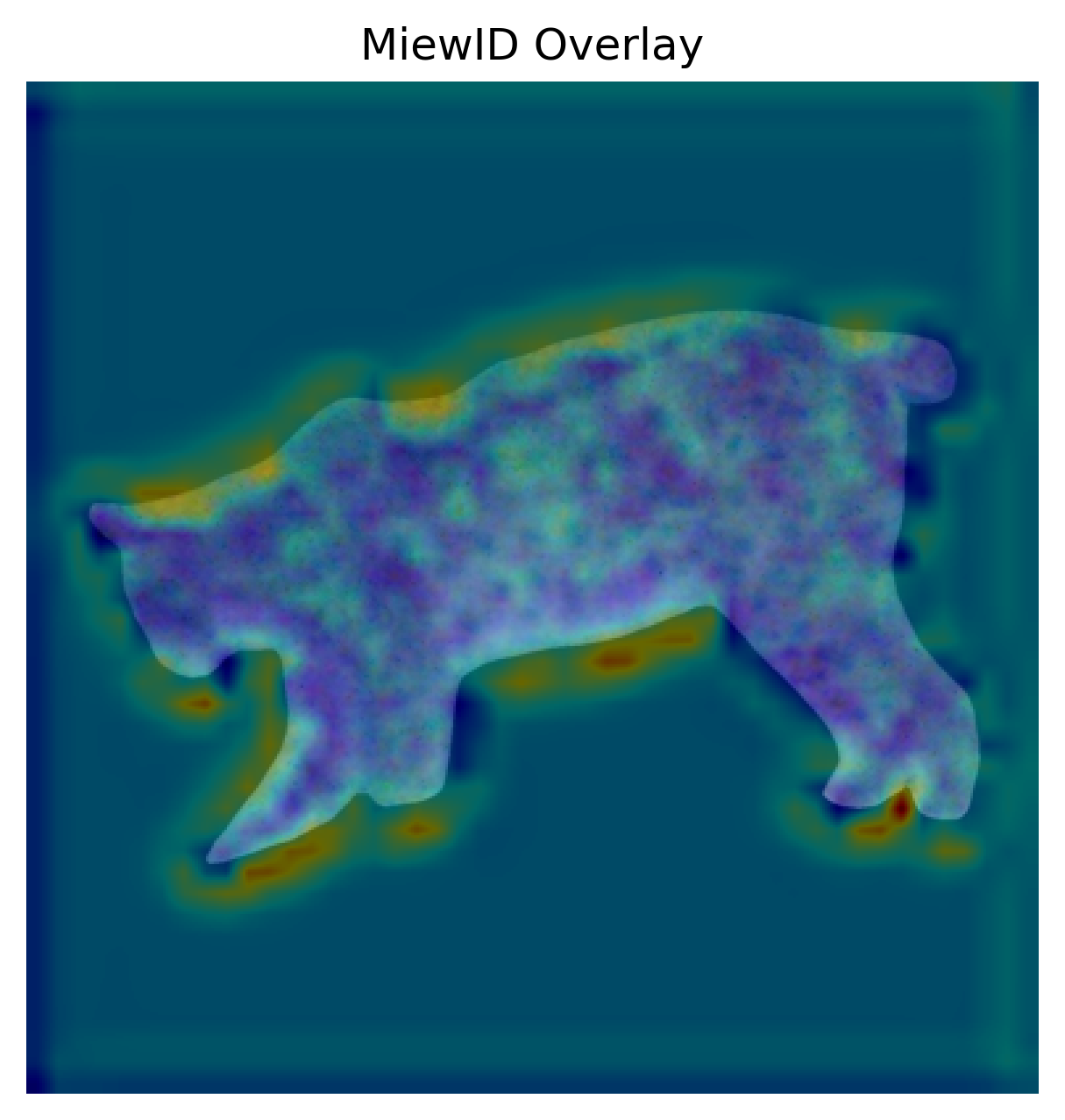} &
\scorecamimg{1.25cm}{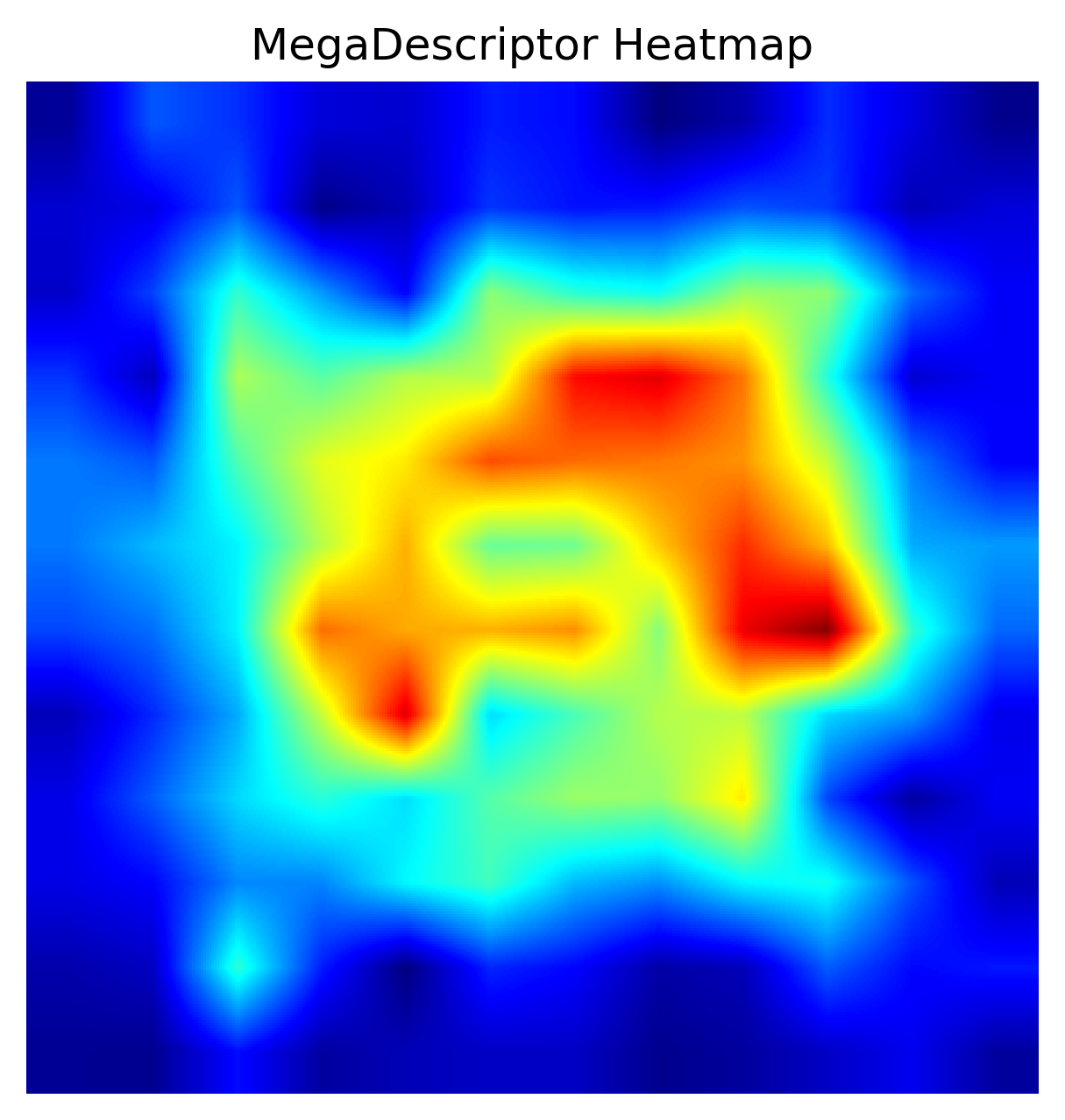} &
\scorecamimg{1.25cm}{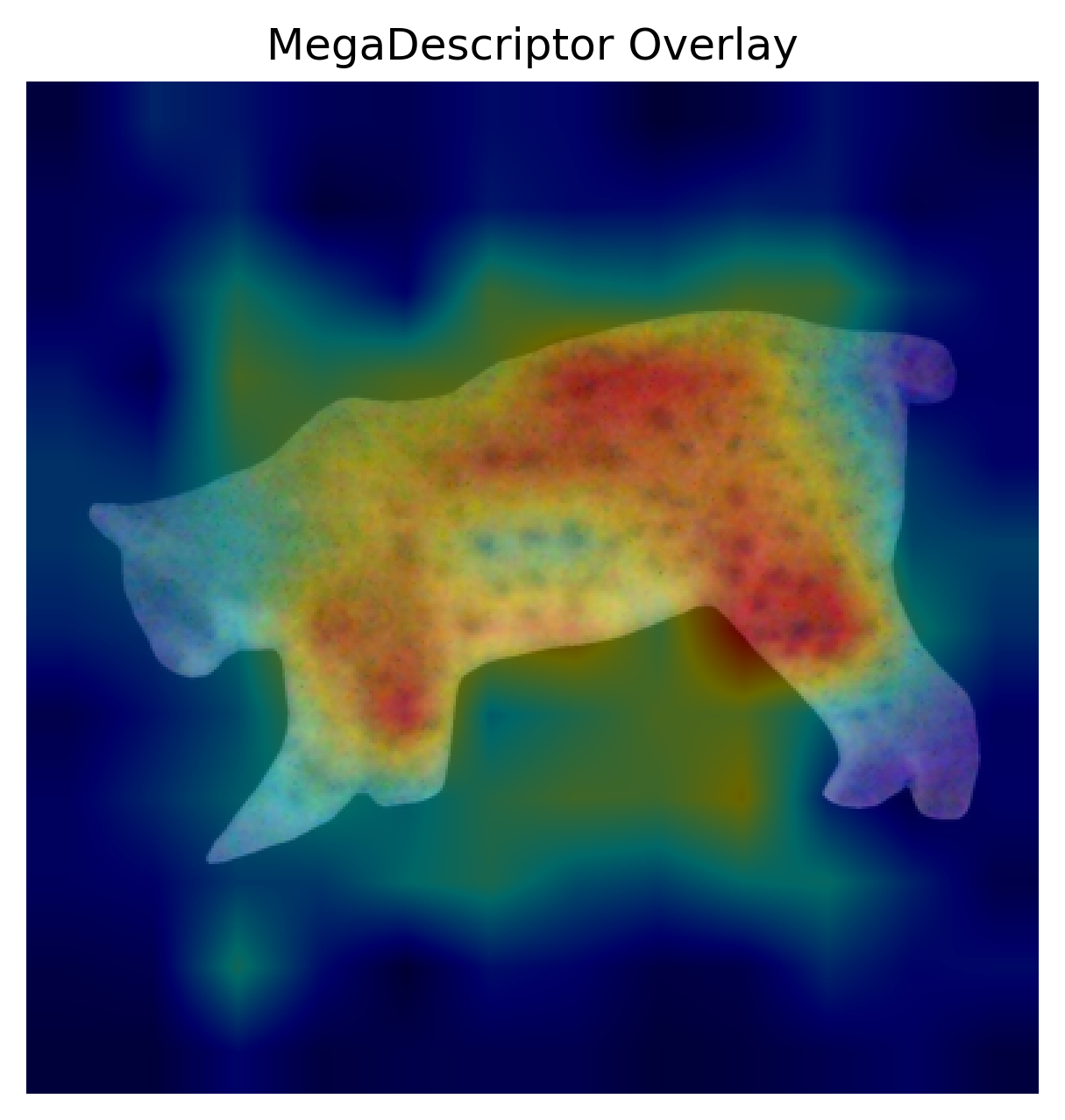}\\[0pt]

{\tiny Salamander} &
\scorecamimg{1.25cm}{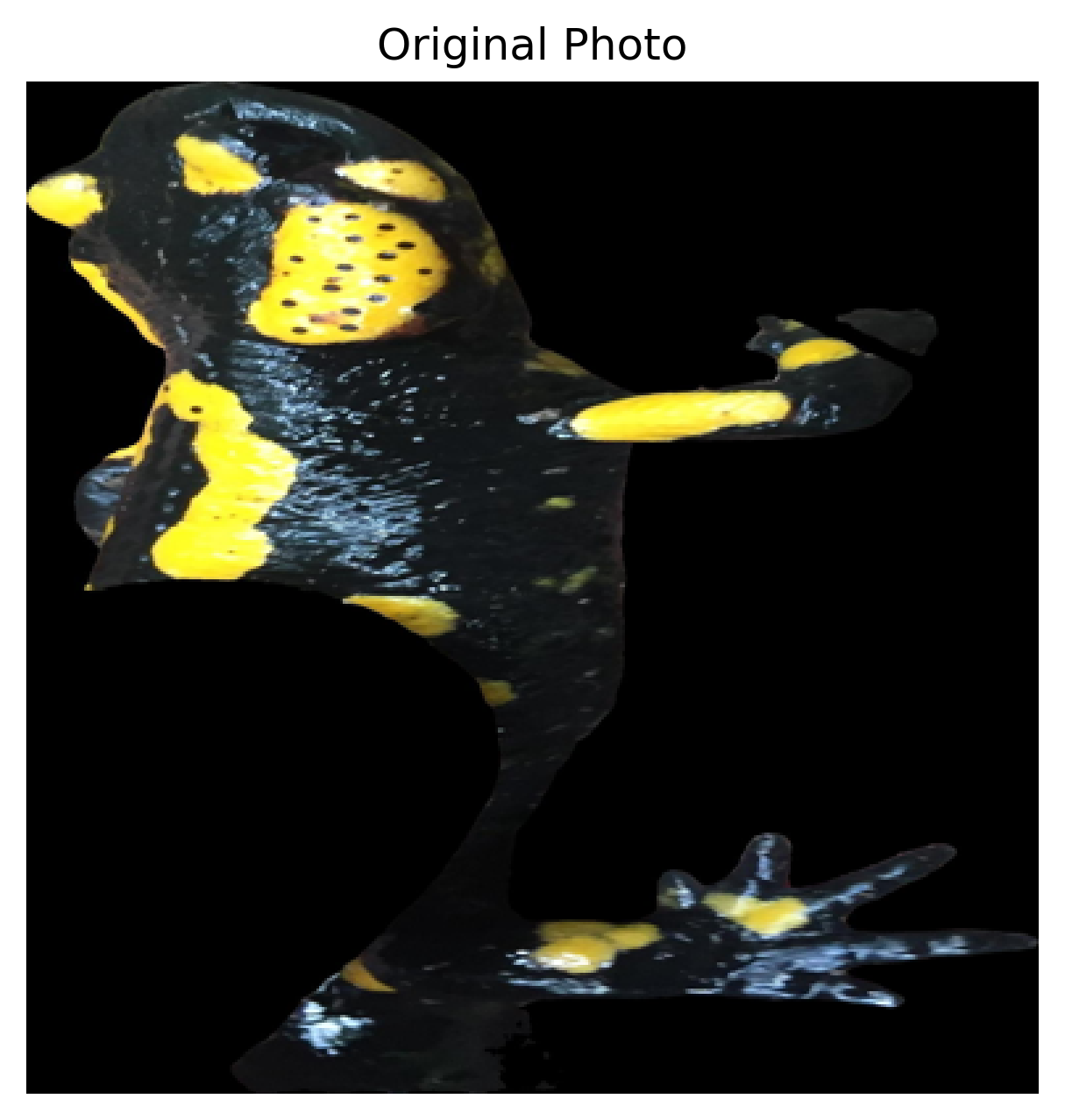} &
\scorecamimg{1.25cm}{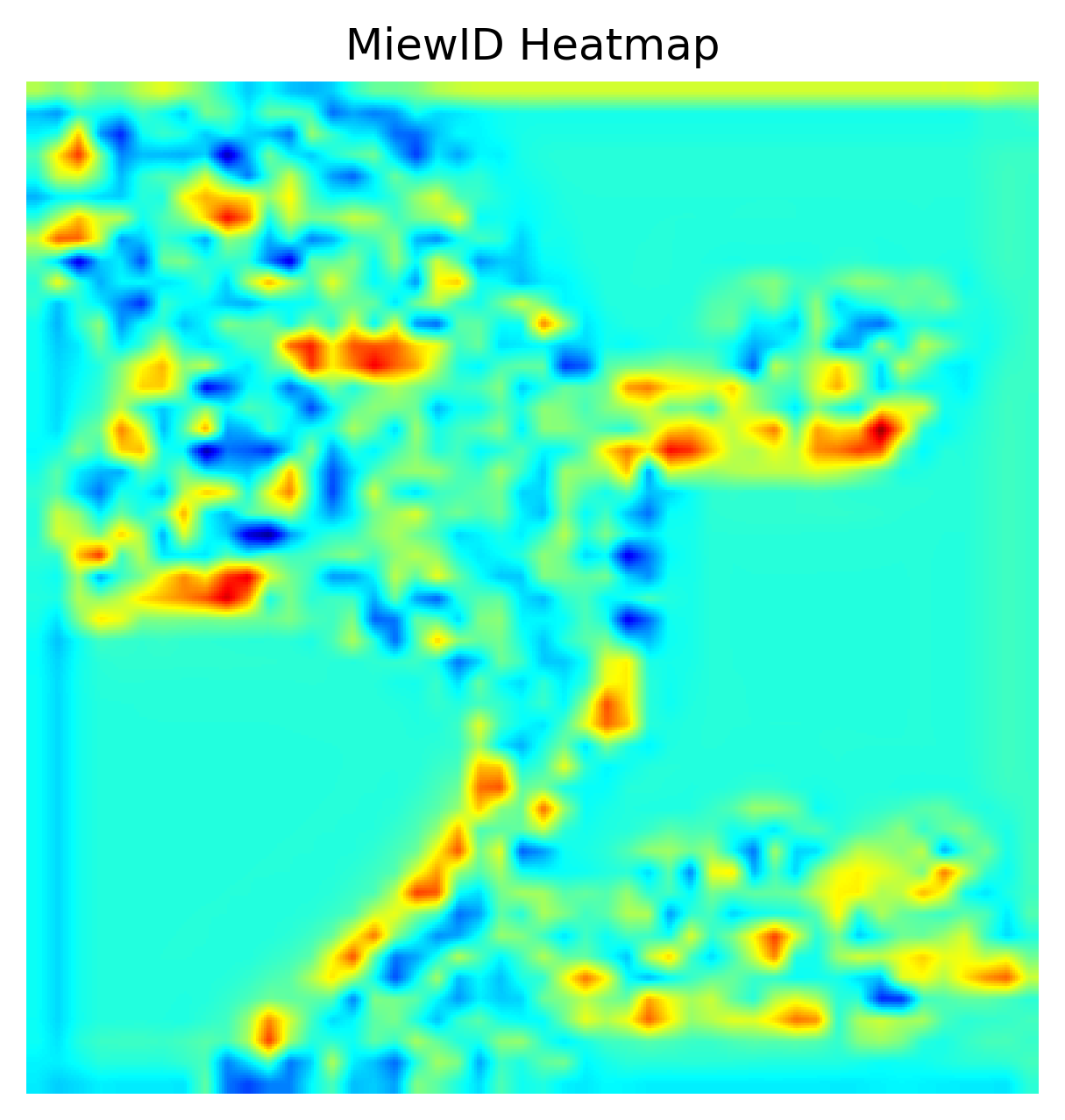} &
\scorecamimg{1.25cm}{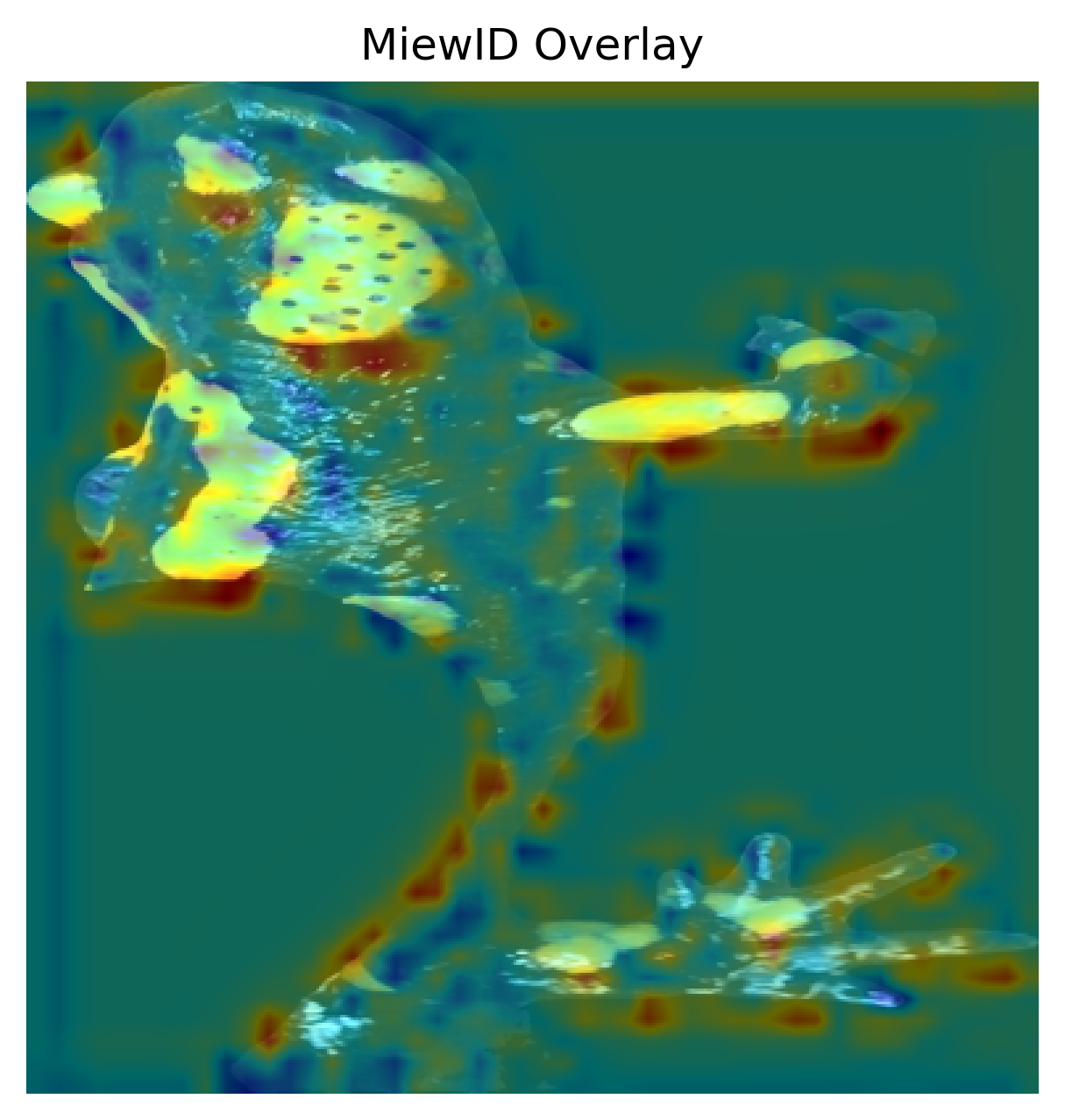} &
\scorecamimg{1.25cm}{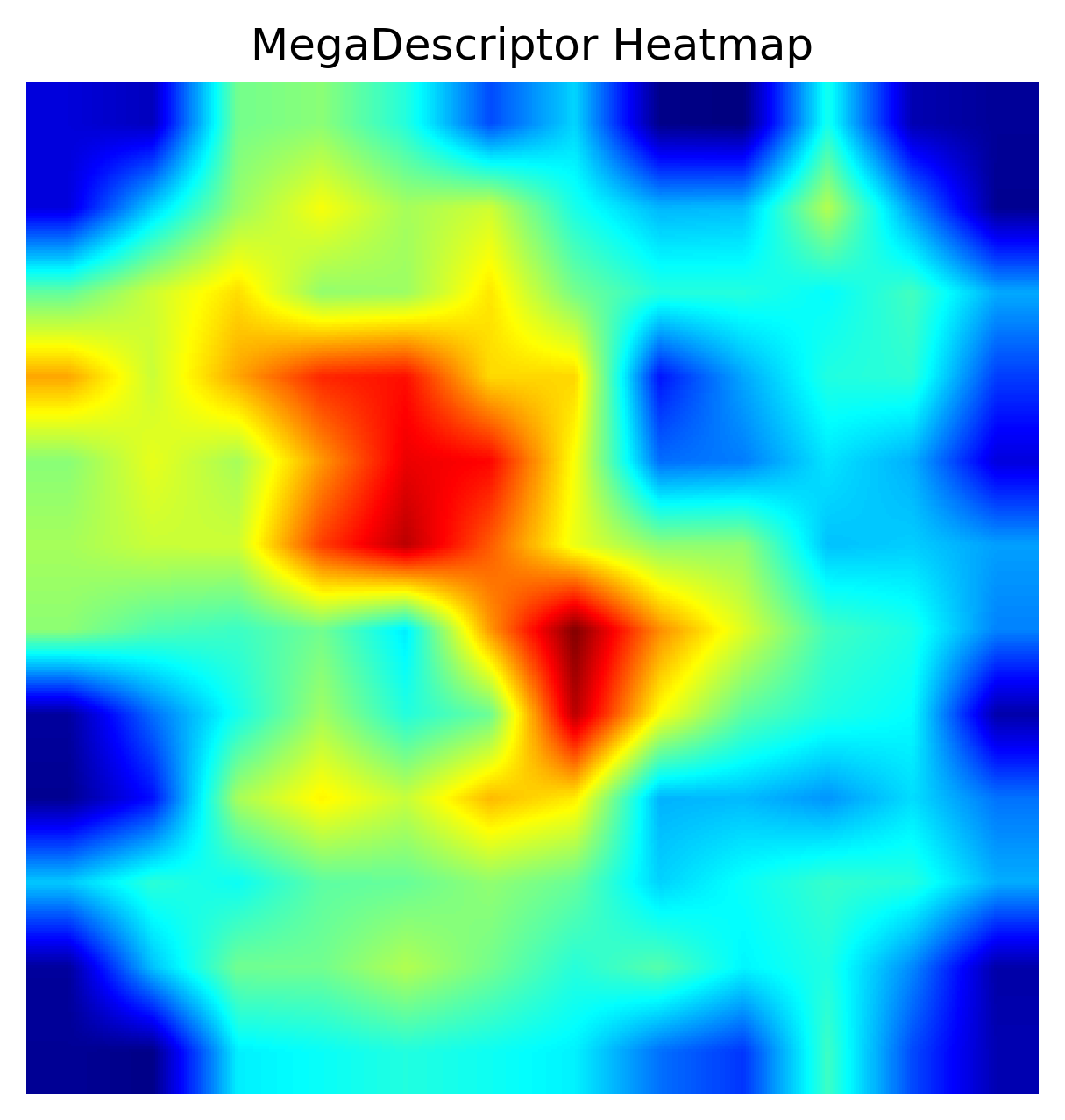} &
\scorecamimg{1.25cm}{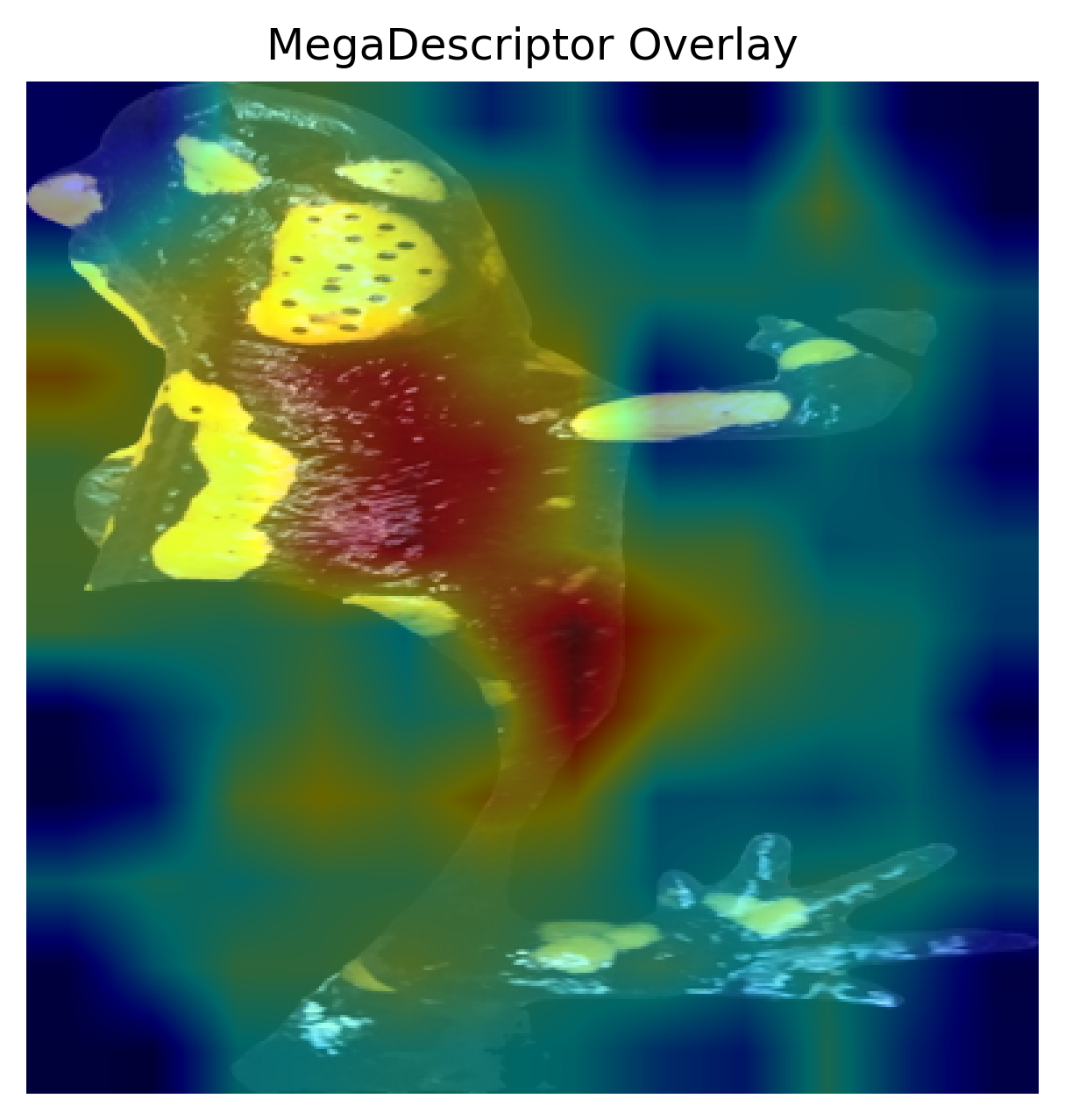}\\[0pt]

{\tiny Sea Turtle} &
\scorecamimg{1.25cm}{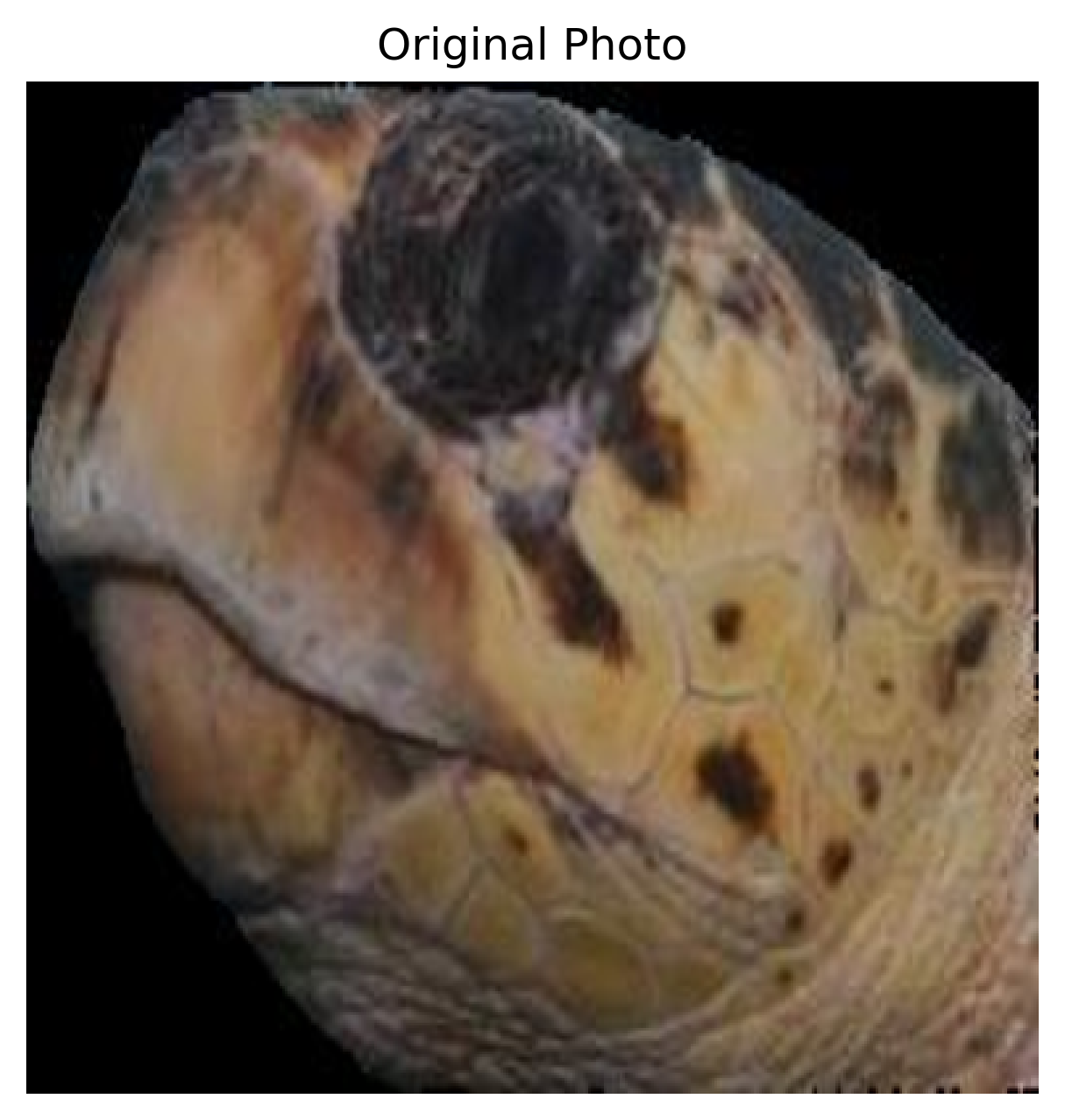} &
\scorecamimg{1.25cm}{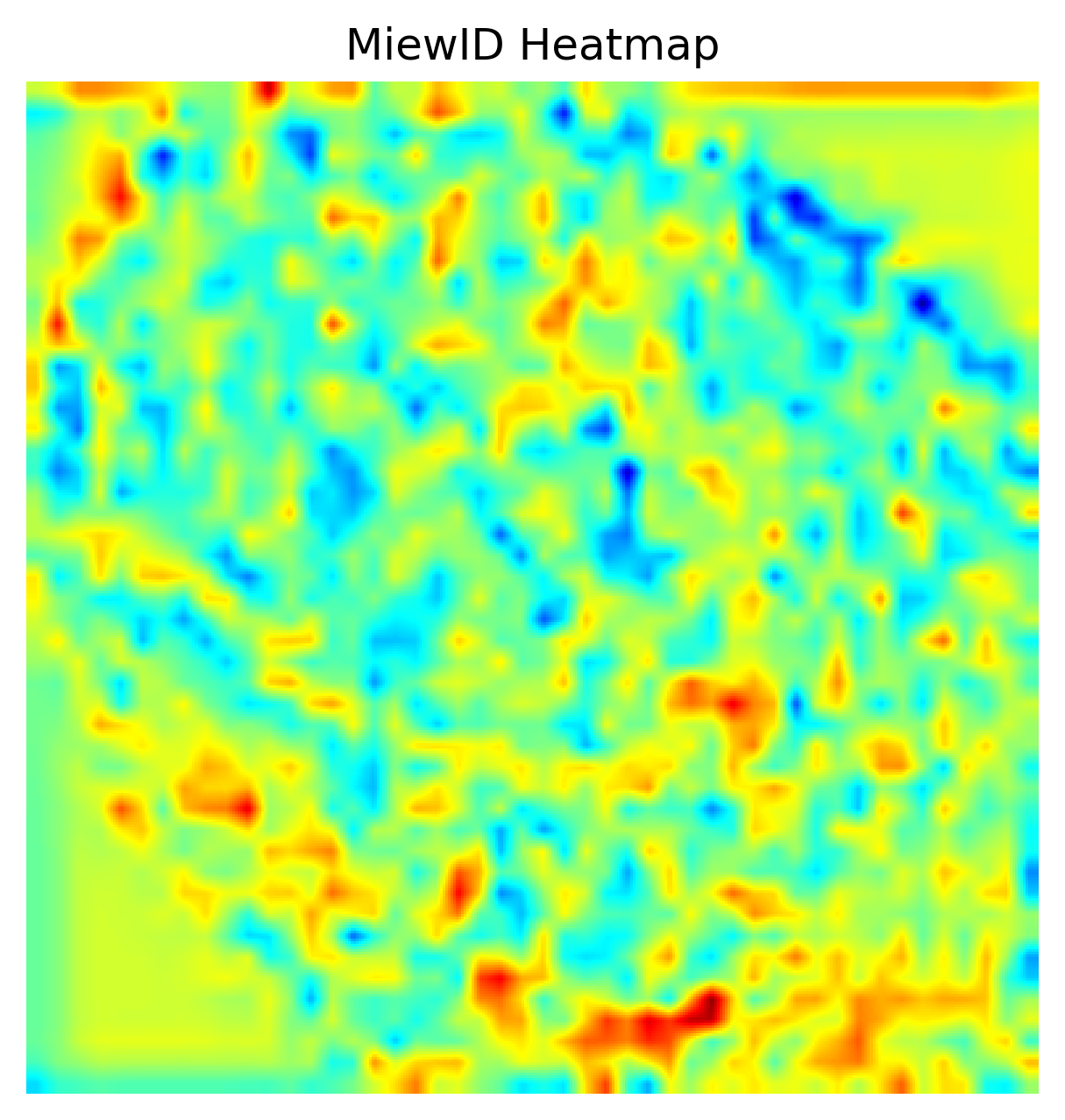} &
\scorecamimg{1.25cm}{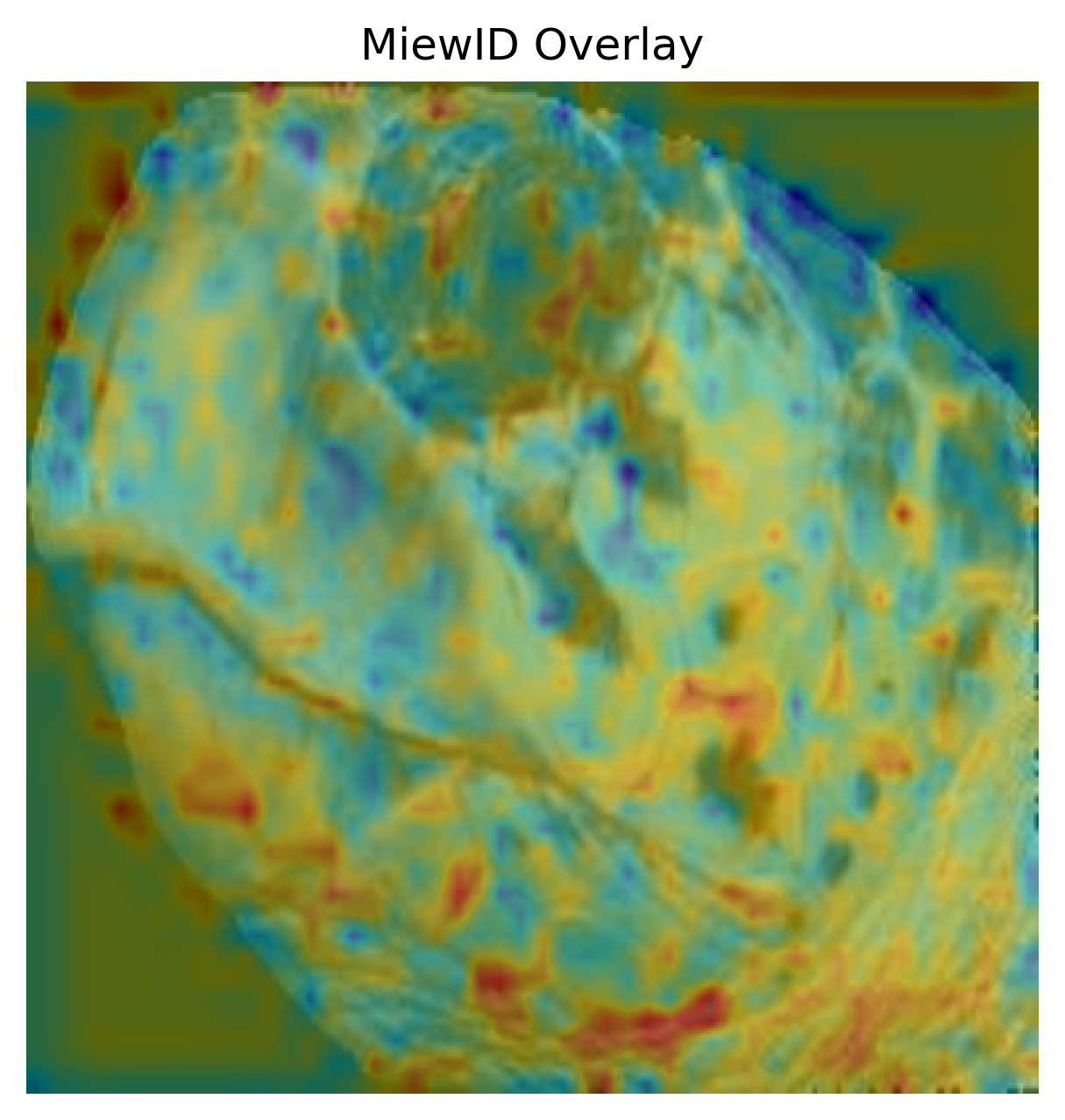} &
\scorecamimg{1.25cm}{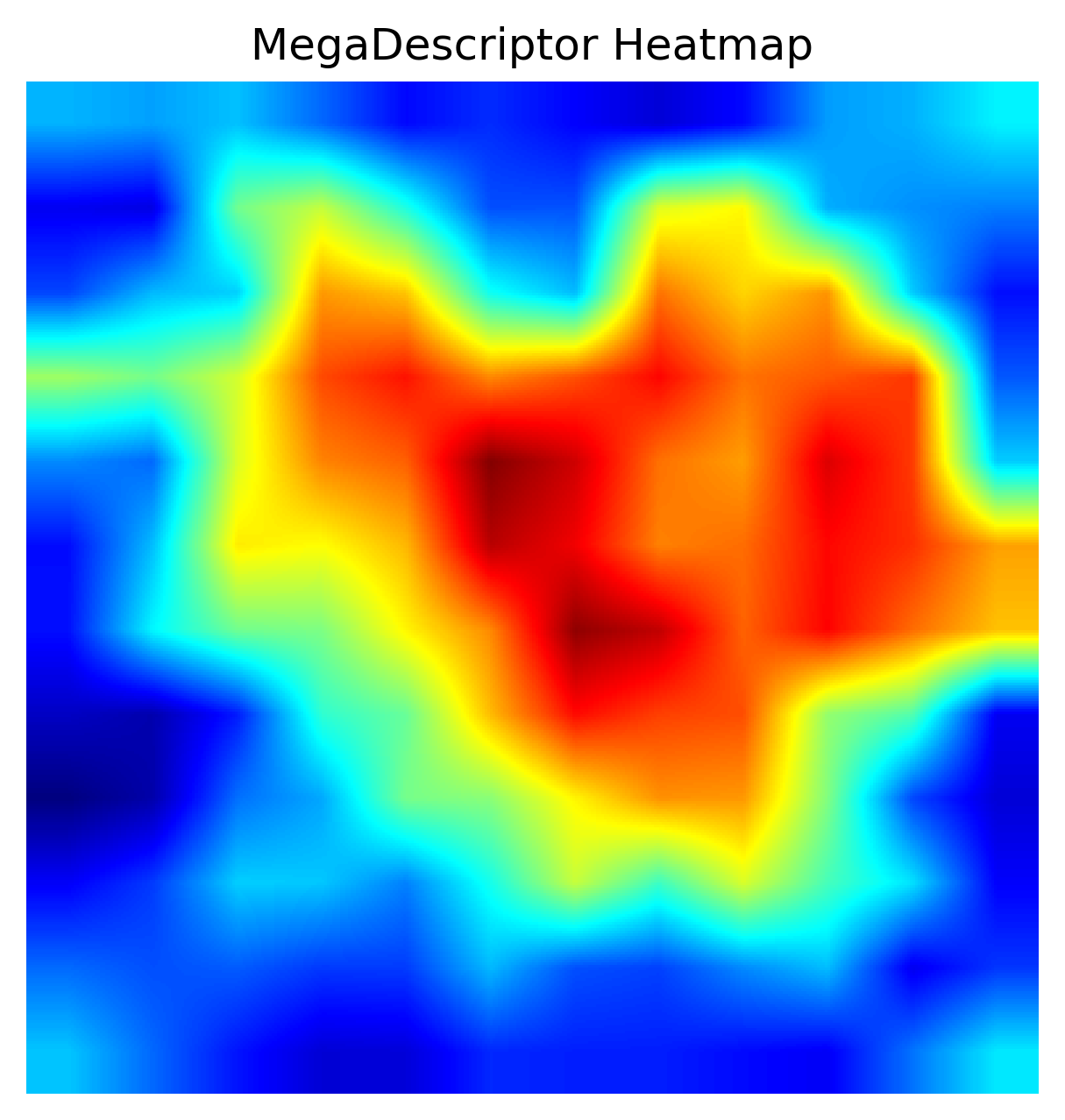} &
\scorecamimg{1.25cm}{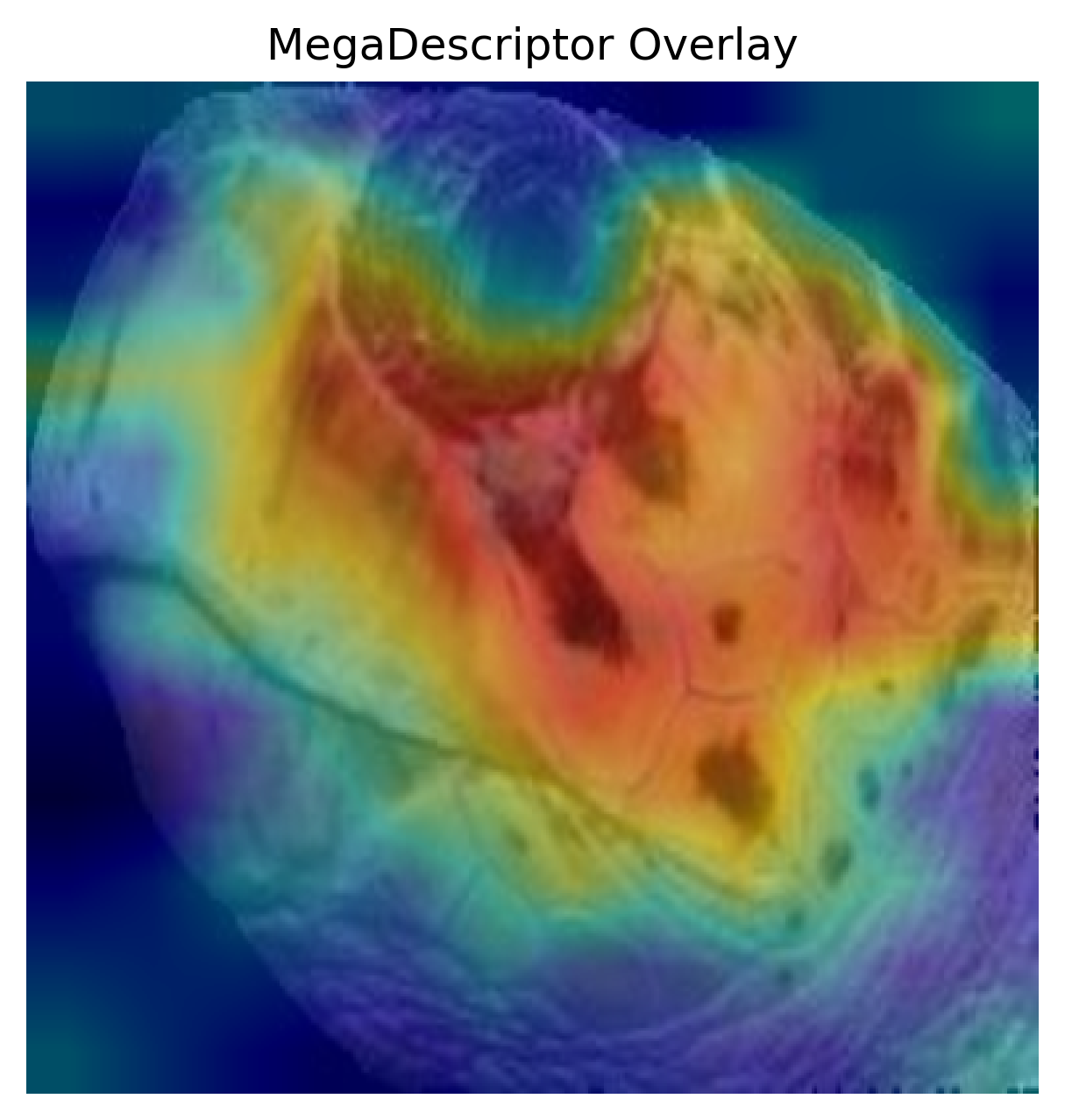}\\[0pt]

{\tiny Horned Lizard} &
\scorecamimg{1.25cm}{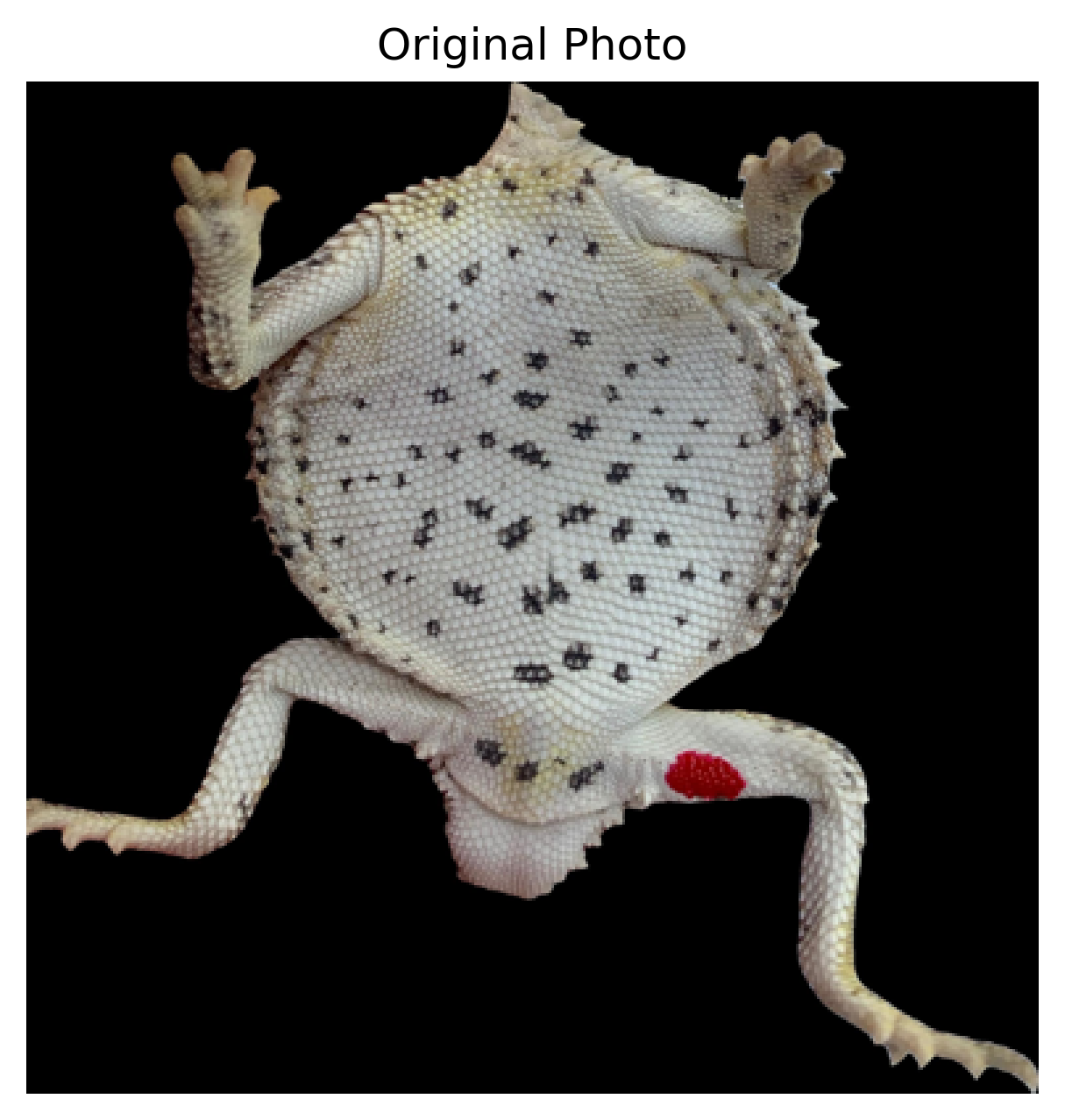} &
\scorecamimg{1.25cm}{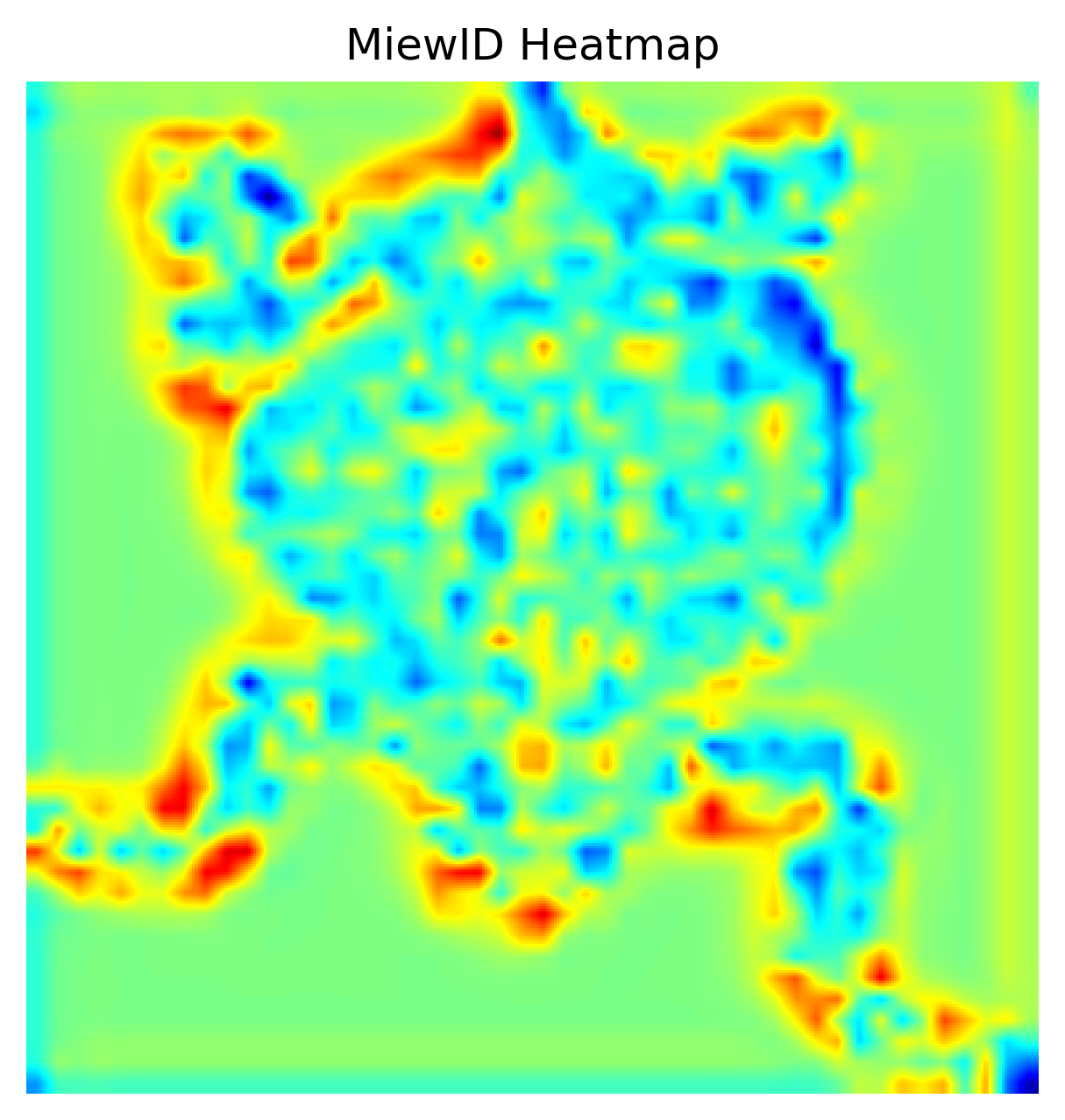} &
\scorecamimg{1.25cm}{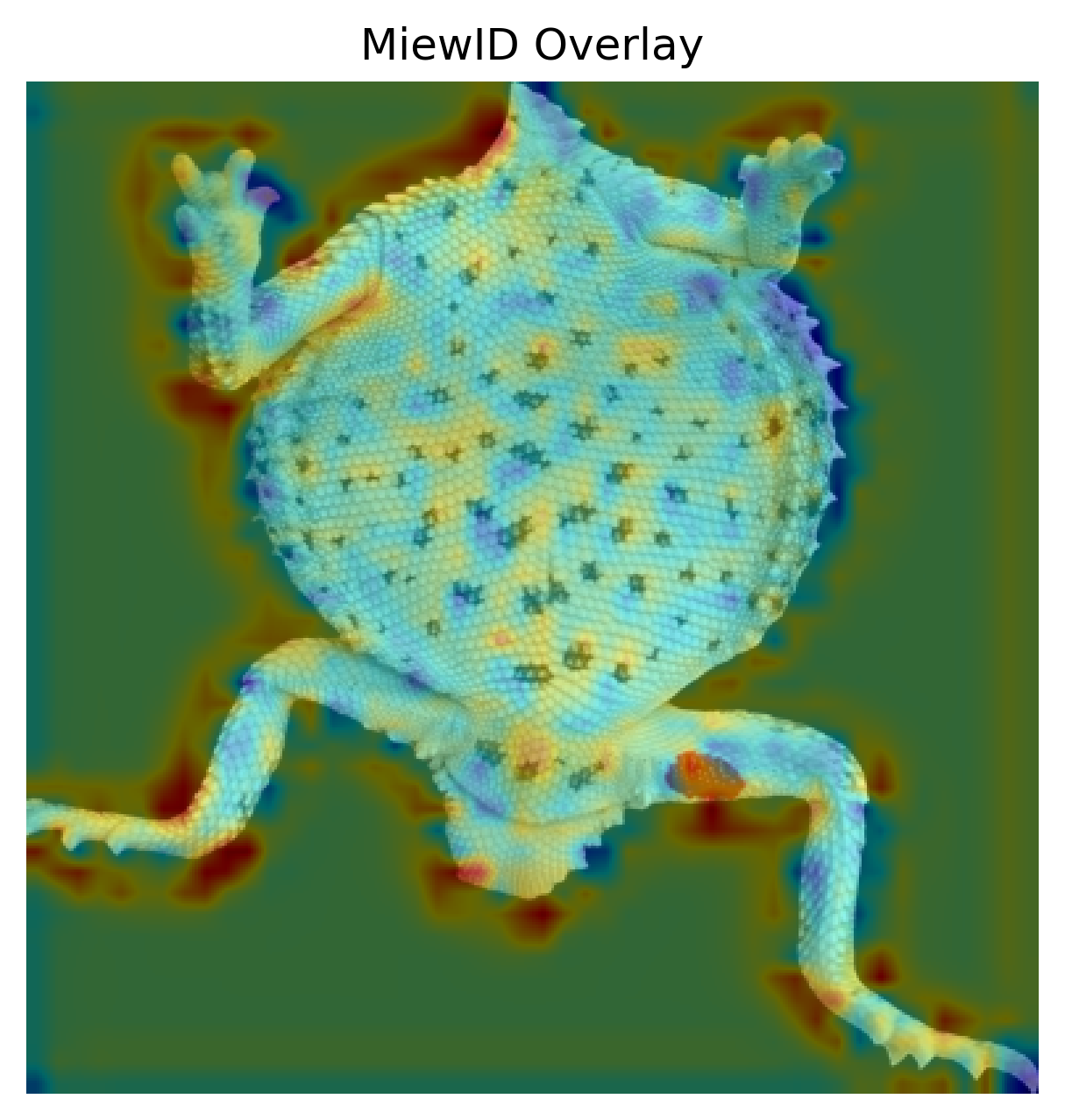} &
\scorecamimg{1.25cm}{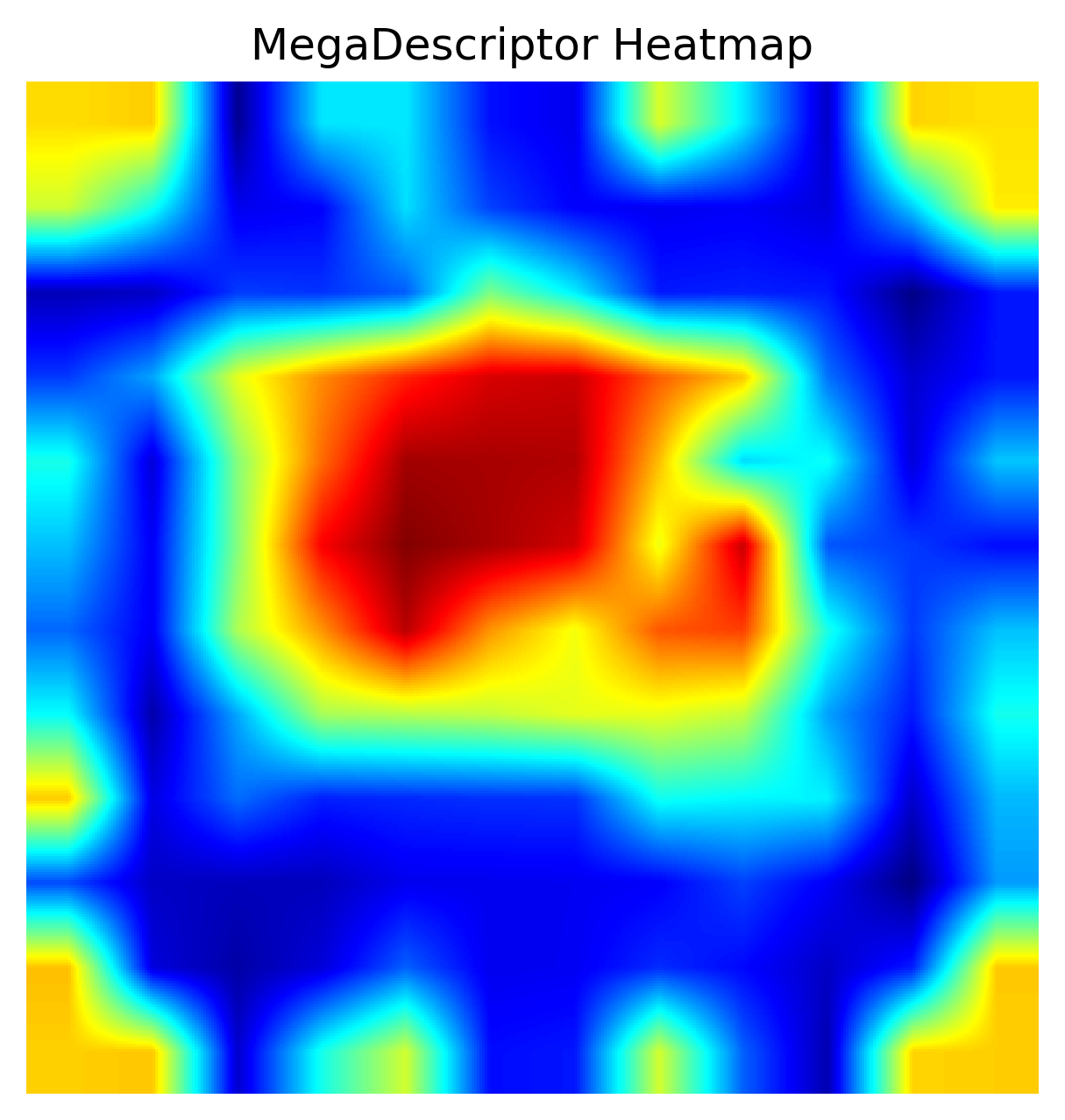} &
\scorecamimg{1.25cm}{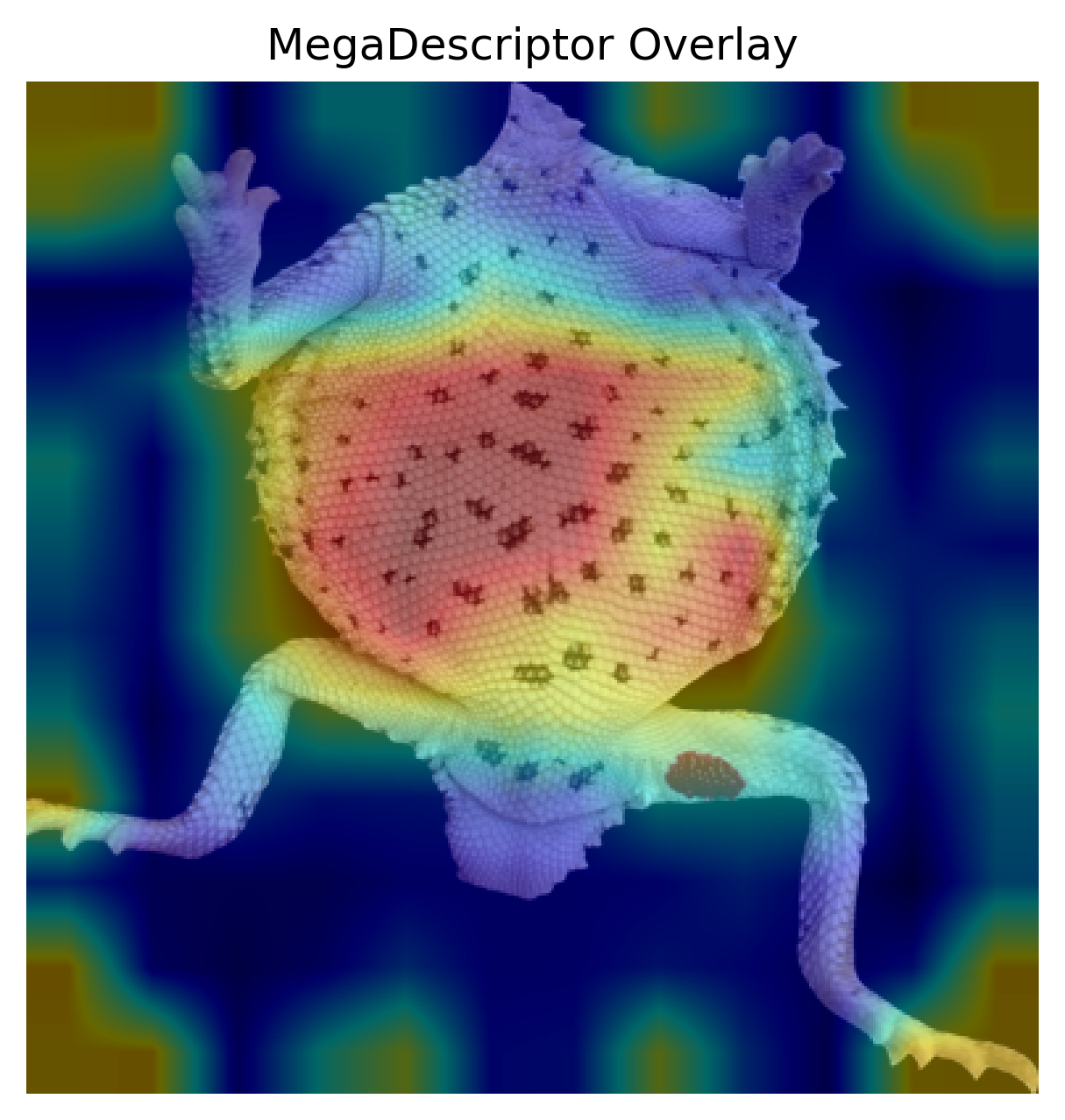}

\end{tabular}%
}

\caption{Score-CAM comparison of MiewID and MegaDescriptor across four species.}
\label{fig:comparison_results}

\end{figure*}

We also considered MegaDescriptor, the foundation model released with the WildlifeDatasets ecosystem \cite{vcermak2024wildlifedatasets}. However, based on qualitative Score-CAM inspection after segmentation, MiewID appeared more suitable in our setting because its heatmaps more often focused on identity-relevant animal regions. In contrast, MegaDescriptor appeared to rely more on broader image-level cues. As shown in Figure~\ref{fig:comparison_results}, each row corresponds to one AnimalCLEF26 species and compares the original image with the MiewID and MegaDescriptor heatmaps and overlays \cite{wang2019scorecam, otarashvili2024multispecies, vcermak2024wildlifedatasets, semenova2025meta}.

To better visualize identity separation, Panel~(a) of Figure~\ref{fig:embedding_grid} shows a UMAP projection of the embedding space produced by the original \texttt{MiewID-msv3} model without fine-tuning, generated with HyperView software~\cite{mcinnes2018umap, hyperview2026}.

\subsubsection{Local Matching Branches}

Local feature matching complemented the MiewID global descriptor by comparing identity-specific anatomical patterns at the correspondence level. Each branch processed image pairs resized to $512 \times 512$ and produced a similarity signal from the number and confidence of matched local features.

\paragraph{ALIKED + LightGlue.}
The first branch used pretrained ALIKED features with LightGlue matching, without further fine-tuning. ALIKED provides efficient keypoint detection and deformable descriptor construction, while LightGlue establishes fast adaptive correspondences between extracted features \cite{zhao2023alikedlighterkeypointdescriptor,lindenberger2023lightgluelocalfeaturematching}.

\paragraph{DISK + LightGlue.}
The second branch used pretrained DISK features with the same LightGlue matcher. DISK is optimized for reliable keypoint detection and description under geometric and photometric variation \cite{tyszkiewicz2020disk}. We retained only matches with confidence at least $0.9$, reducing ambiguous correspondences and providing a complementary local descriptor family.

\begin{figure}[!htbp]
    \centering
    \scriptsize
    \setlength{\tabcolsep}{2pt}
    \renewcommand{\arraystretch}{0.85}
    \begin{tabular}{@{}ccc@{}}
        \textbf{LynxID2025} & \textbf{SalamanderID2025} & \textbf{SeaTurtleID2022} \\
        \includegraphics[width=0.30\linewidth]{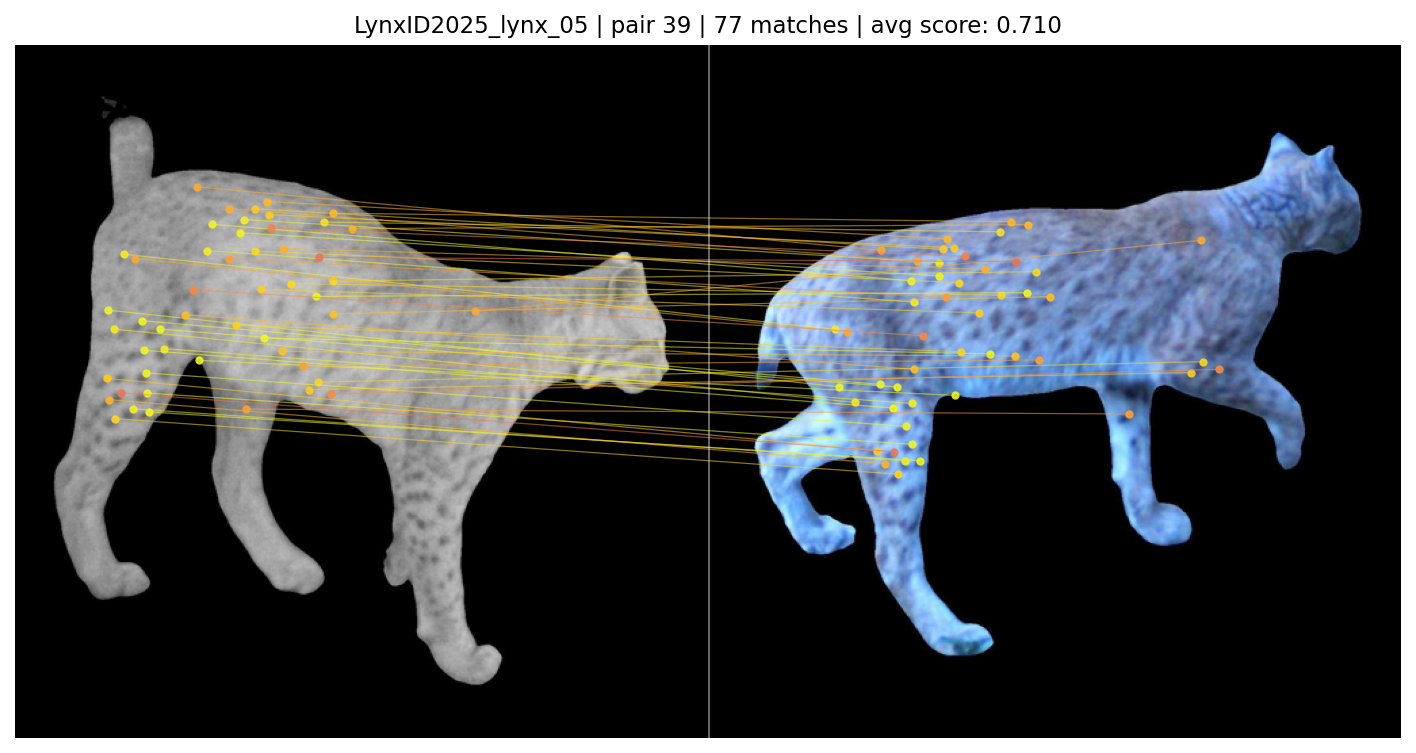} &
        \includegraphics[width=0.30\linewidth]{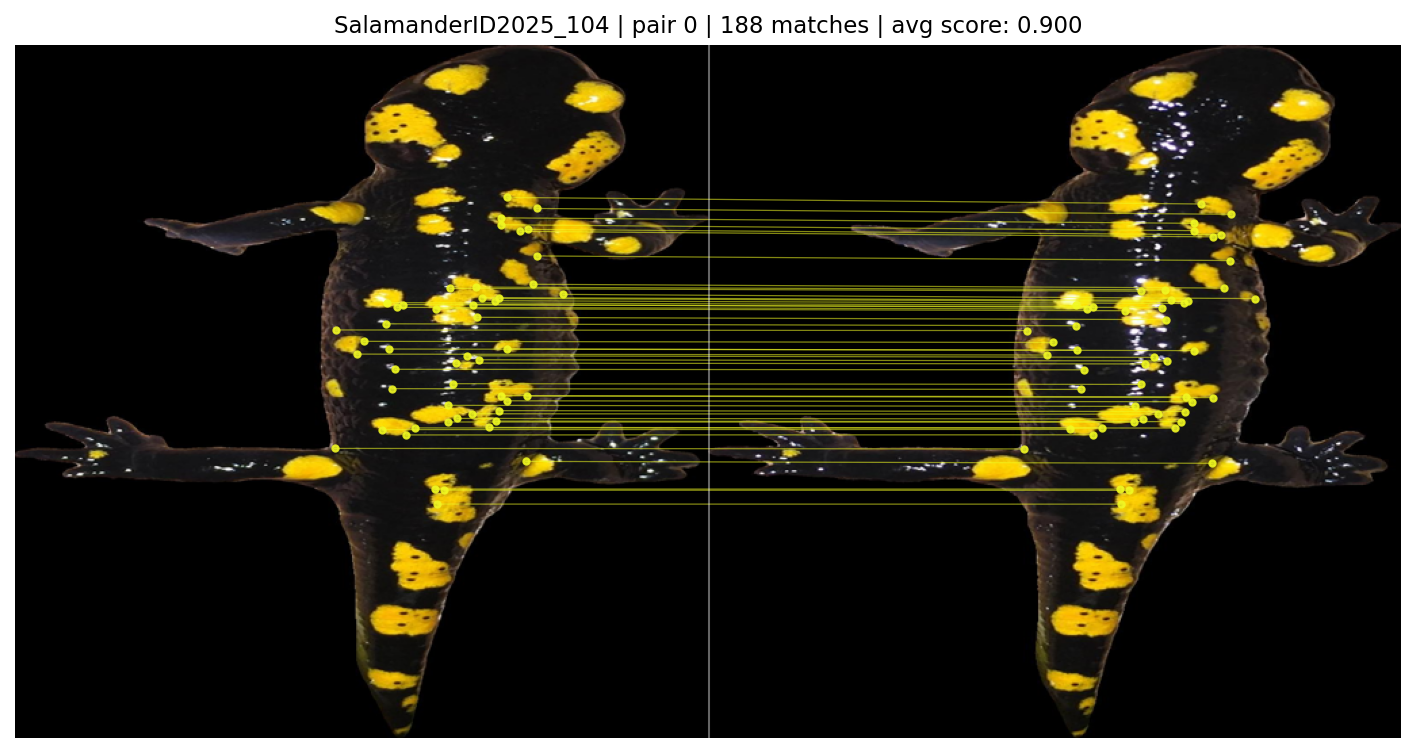} &
        \includegraphics[width=0.30\linewidth]{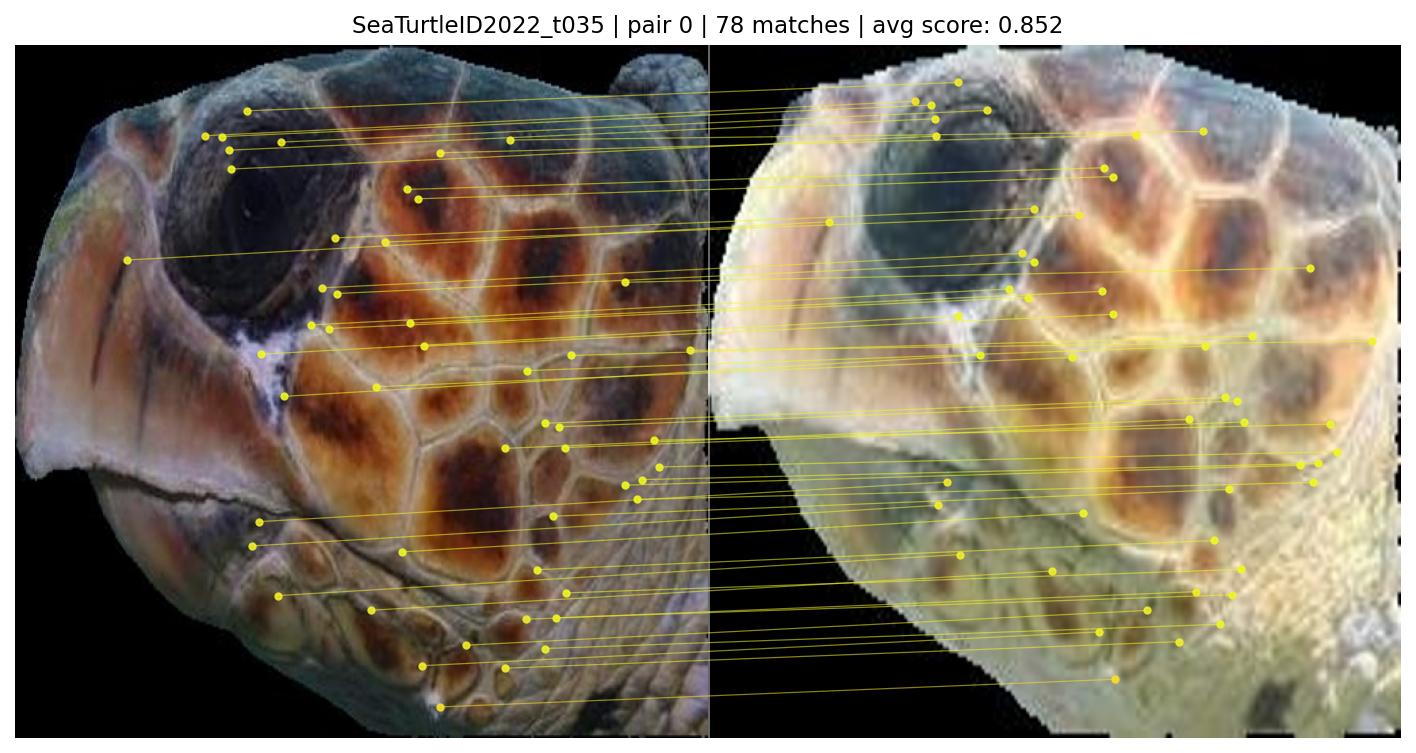} \\
        \multicolumn{3}{c}{\tiny ALIKED + LightGlue} \\[1pt]
        \includegraphics[width=0.30\linewidth]{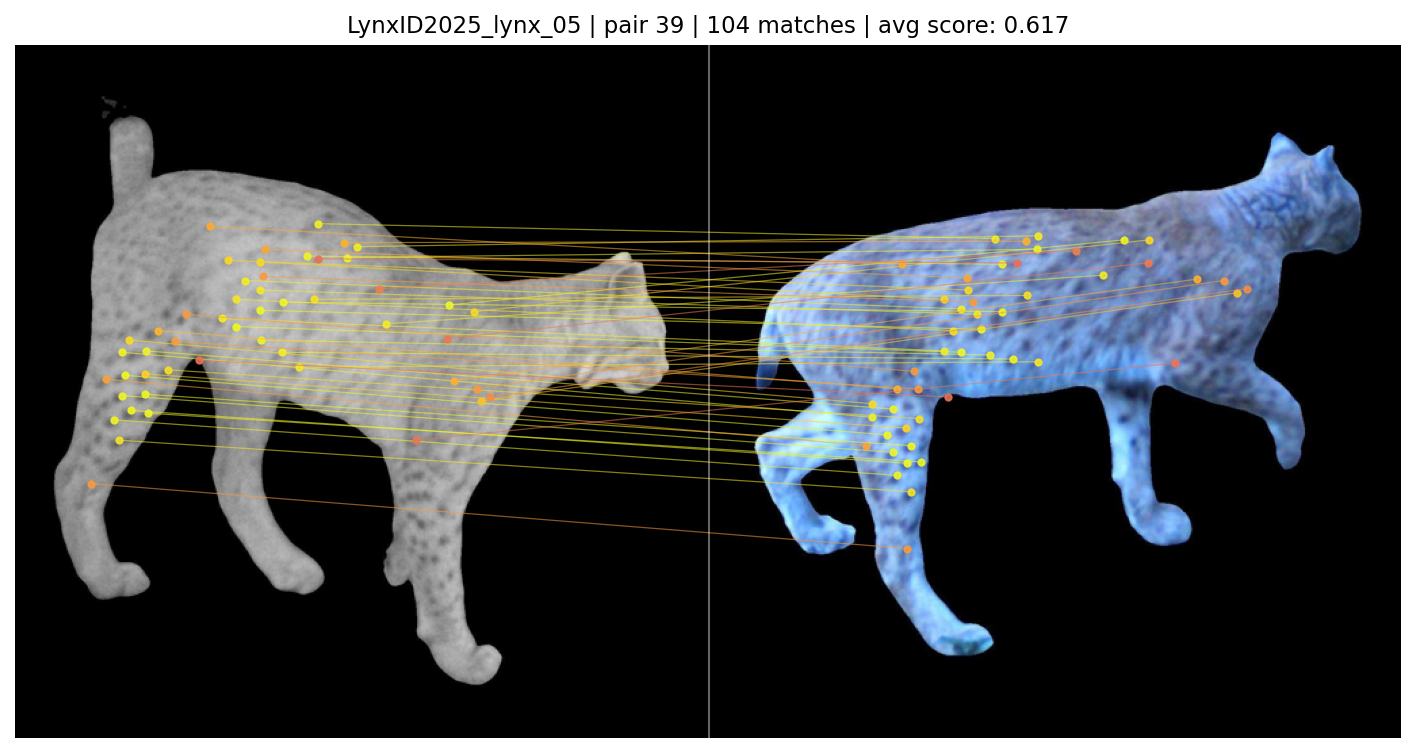} &
        \includegraphics[width=0.30\linewidth]{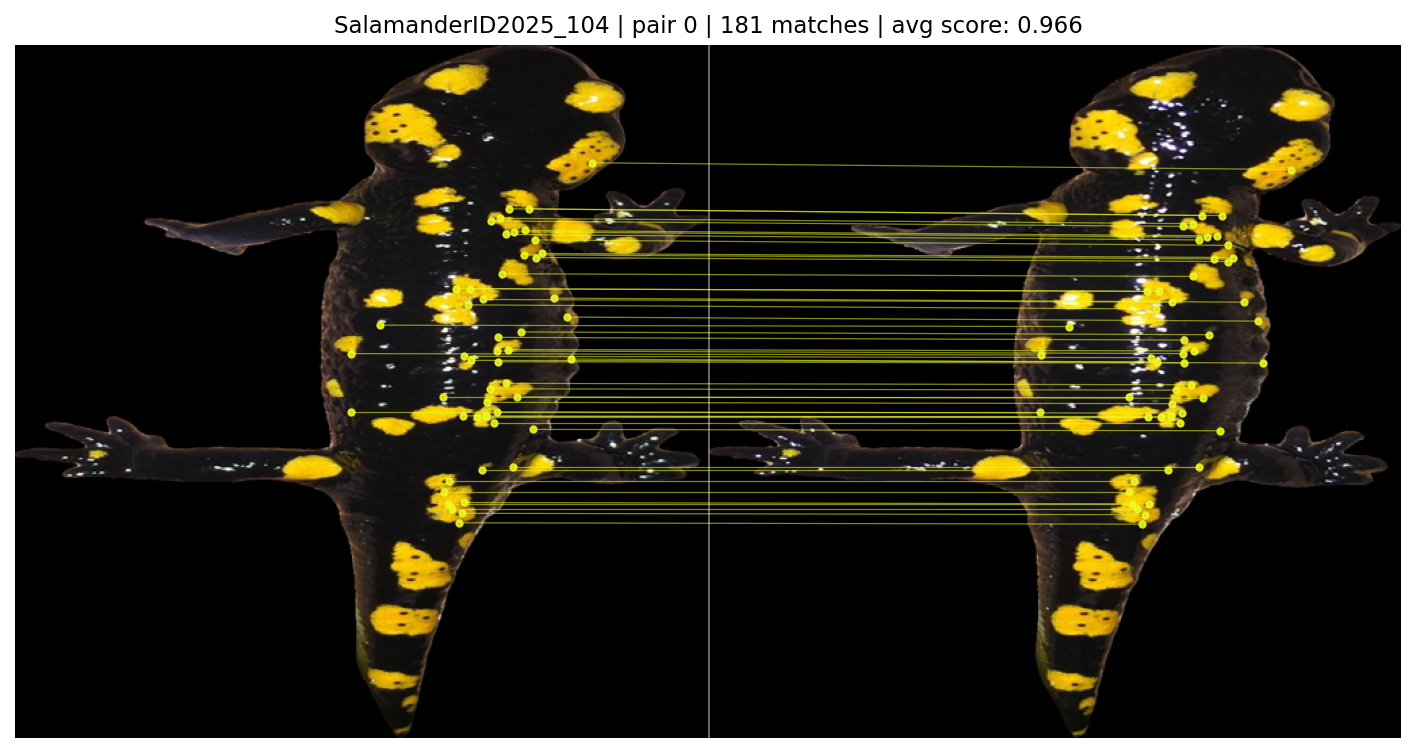} &
        \includegraphics[width=0.30\linewidth]{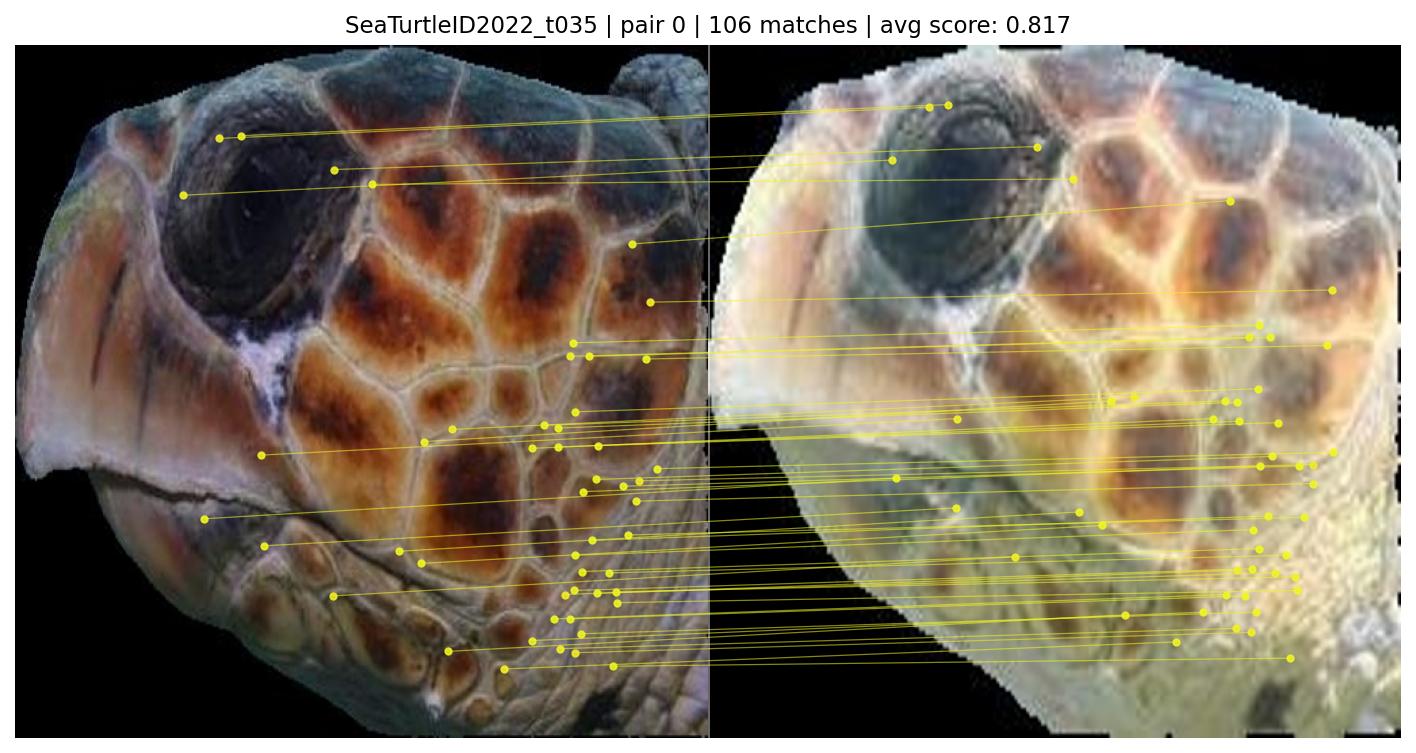} \\
        \multicolumn{3}{c}{\tiny DISK + LightGlue}
    \end{tabular}
    \caption{Local matcher comparison across three datasets}
    \label{fig:local_matchers_comparison}
\end{figure}

\paragraph{Qualitative Analysis of Local Matchers.}
Figure~\ref{fig:local_matchers_comparison} compares ALIKED + LightGlue and DISK + LightGlue across lynx, salamander, and sea turtle examples. ALIKED tends to produce structured keypoints around salient contours and repeated markings, whereas DISK often yields denser correspondences over textured regions. This complementary behavior motivated using both branches in WildFusion rather than relying on a single local descriptor.

\FloatBarrier

\subsection{WildFusion Similarity Computation and Calibration}

Using only global descriptors could miss fine local correspondences, while relying only on local matching was often computationally expensive and less stable at scale. To address this trade-off, we built on WildFusion, a calibrated similarity-fusion framework for individual animal identification \cite{cermak2024wildfusion}. WildFusion combines global descriptor similarity with local matching evidence into a unified score, allowing the system to benefit from both holistic appearance cues and precise local correspondences.

In our setup, WildFusion used three similarity branches: the MiewID global descriptor, ALIKED + LightGlue, and DISK + LightGlue. We used \texttt{IsotonicCalibration} for score calibration, with the WildFusion parameter $B$ set to $256$. The calibrated branch scores were then fused using the standard WildFusion averaging rule.

This design was well suited to our discovery-oriented setting. Global embeddings provided robust coarse identity structure, whereas local matches helped disambiguate visually similar individuals. By integrating both signals, WildFusion offered a stronger basis for downstream clustering and identity assignment than either component alone.

\subsection{Clustering and Known-Identity Matching}

After computing calibrated WildFusion similarities, our final stage converted pairwise image similarities into submission labels. The procedure was applied independently for each species, allowing the similarity refinement, graph construction, clustering, and known-identity attachment parameters to be selected separately for each dataset. For a given species, let $\mathcal{Q}=\{q_i\}_{i=1}^{N}$ denote the query images and let $\mathcal{G}=\{g_j\}_{j=1}^{M}$ denote the corresponding database images, when a database was available.

We computed two similarity matrices using WildFusion. The query--query matrix
\[
S^{QQ} \in [0,1]^{N \times N}
\]
was used for clustering query images into individual animals. When database images were available, we also computed the query--database matrix
\[
S^{QG} \in [0,1]^{N \times M},
\]
which was used only for known-identity attachment. Both matrices were clipped to the range $[0,1]$ for numerical stability. Importantly, these two matrices were used for different purposes: $S^{QQ}$ defined the graph structure for query clustering, whereas $S^{QG}$ provided evidence for assigning clusters to known database identities.

\subsubsection{K-Reciprocal Re-Ranking of Query--Query Similarities}

When enabled for a given species, we applied k-reciprocal re-ranking to the query--query similarity matrix before graph clustering \cite{zhong2017re}. Re-ranking was restricted to $S^{QQ}$; the query--database matrix $S^{QG}$ used for known-identity attachment was left unchanged. This separation prevented the neighborhood refinement used for clustering from altering the calibrated query--database scores used to decide whether a query belonged to a known identity.

Let $S^{\mathrm{orig}} = S^{QQ}$ denote the original WildFusion query--query similarity matrix, and define the corresponding distance matrix as
\[
d^{\mathrm{orig}}_{ij}=1-S^{\mathrm{orig}}_{ij}.
\]
For each query image $q_i$, we retrieved the $k_1$ nearest neighbors according to $d^{\mathrm{orig}}$. A neighbor $q_j$ was retained in the reciprocal set of $q_i$ only if the relation was mutual: $q_j$ had to be among the top-$k_1$ neighbors of $q_i$, and $q_i$ also had to be among the top-$k_1$ neighbors of $q_j$. This gave the initial k-reciprocal set
\[
\mathcal{R}_{k_1}(i)
=
\{j \mid j \in \mathcal{N}_{k_1}(i),\; i \in \mathcal{N}_{k_1}(j)\}.
\]

The reciprocal set was then expanded to include reliable neighbors that might be missed by the strict mutual-neighbor rule. For each $j \in \mathcal{R}_{k_1}(i)$, we computed a smaller reciprocal set $\mathcal{R}_{\lceil k_1/2\rceil}(j)$. This set was absorbed into the neighborhood of $q_i$ when more than two-thirds of its members were already present in $\mathcal{R}_{k_1}(i)$:
\[
\frac{
|\mathcal{R}_{\lceil k_1/2\rceil}(j) \cap \mathcal{R}_{k_1}(i)|
}{
|\mathcal{R}_{\lceil k_1/2\rceil}(j)|
}
>
\frac{2}{3}.
\]
We denoted the expanded reciprocal set by $\mathcal{R}^{*}_{k_1}(i)$.

Each expanded set was encoded as a sparse affinity vector $v_i$, where closer neighbors received larger weights:
\[
v_i(j)
=
\frac{\exp(-d^{\mathrm{orig}}_{ij})}
{\sum_{t \in \mathcal{R}^{*}_{k_1}(i)}
\exp(-d^{\mathrm{orig}}_{it})},
\quad
j \in \mathcal{R}^{*}_{k_1}(i),
\]
and $v_i(j)=0$ otherwise. When $k_2>1$, this vector was smoothed by averaging it over the $k_2$ nearest neighbors in the original ranking. This reduced the effect of unstable individual neighbor assignments and produced a more robust local neighborhood representation.

A continuous Jaccard distance was then computed between the re-ranked neighborhood vectors:
\[
d^{J}_{ij}
=
1 -
\frac{
\sum_l \min(v_i(l), v_j(l))
}{
\sum_l \max(v_i(l), v_j(l))
}.
\]
The final distance combined this neighborhood-based distance with the original WildFusion distance:
\[
d^{\mathrm{final}}_{ij}
=
(1-\lambda)d^{J}_{ij}
+
\lambda d^{\mathrm{orig}}_{ij},
\]
where $\lambda \in [0,1]$ controlled the contribution of the original pairwise distance. The re-ranked similarity matrix was then obtained as
\[
S^{\mathrm{rr}}_{ij}=1-d^{\mathrm{final}}_{ij}.
\]

The matrix passed to clustering was therefore
\[
S^{\mathrm{CW}}
=
\begin{cases}
S^{\mathrm{rr}}, & \text{if re-ranking is enabled for the species},\\
S^{QQ}, & \text{otherwise}.
\end{cases}
\]
The re-ranking parameters $(k_1,k_2,\lambda)$ were selected as part of the species-specific clustering configuration, together with the graph-construction and label-propagation hyperparameters.

\subsubsection{Graph Construction and Chinese Whispers Clustering}

The selected query--query similarity matrix $S^{\mathrm{CW}}$ was converted into a sparse weighted graph. Each query image was represented as a node, and edge weights were derived from the pairwise similarities in $S^{\mathrm{CW}}$. This representation was well suited to the open-set nature of the task because the number of identities in the query set was unknown in advance. Instead of requiring a predefined number of clusters, Chinese Whispers inferred clusters from dense connected regions in the graph \cite{biemann2006chinese}.

Graph construction was controlled by a species-specific similarity threshold $\tau_{\mathrm{CW}}$, nearest-neighbor limit $k_{\mathrm{NN}}$, and k-core degree $k_{\mathrm{core}}$. Edges with similarity below $\tau_{\mathrm{CW}}$ were removed. The graph was then sparsified using nearest-neighbor constraints so that clustering depended mainly on the strongest local relationships rather than weak long-range similarities. A k-core filtering step was used to preserve a reliable graph backbone by removing weakly connected nodes whose degree was below $k_{\mathrm{core}}$. If this filtering removed all nodes, the algorithm fell back to the unfiltered graph to avoid degeneracy.

Chinese Whispers was then applied to the resulting weighted graph. Each node was initialized with a unique label. During each iteration, nodes were visited in randomized order, and each node adopted the label with the largest total incoming edge weight among its neighbors. The process was repeated until convergence or until the maximum number of iterations was reached. The final labels were remapped to contiguous cluster indices and used as the predicted query clusters.

In our implementation, Chinese Whispers received four species-specific hyperparameters:
\[
(\tau_{\mathrm{CW}}, k_{\mathrm{NN}}, k_{\mathrm{core}}, T),
\]
where $\tau_{\mathrm{CW}}$ was the edge threshold, $k_{\mathrm{NN}}$ was the top-neighbor limit, $k_{\mathrm{core}}$ was the minimum backbone degree, and $T$ was the maximum number of label-propagation iterations.

\subsubsection{Known-Identity Scoring}
For species with a labeled database, we used the query--database matrix $S^{QG}$ to estimate whether each query image matched a known identity. Let $\mathcal{G}_c$ denote the set of database images belonging to known identity $c$. For each query image $q_i$ and each known identity $c$, we converted image-level similarities into an identity-level score by averaging the top-$r$ similarities between $q_i$ and the database images of identity $c$:
\[
A_{i,c}
=
\frac{1}{\min(r,|\mathcal{G}_c|)}
\sum_{s \in \operatorname{Top}_{r}
\left(\{S^{QG}_{ij} \mid g_j \in \mathcal{G}_c\}\right)}
s.
\]
The top-$r$ aggregation made the identity score less sensitive to low-quality or viewpoint-mismatched database images, while still allowing multiple strong matches to support a known-identity assignment.

For each query image, we then identified the best and second-best known identities:
\[
c_i^{(1)} = \arg\max_c A_{i,c},
\]
\[
b_i = A_{i,c_i^{(1)}},
\]
\[
b_i^{(2)} = \max_{c \neq c_i^{(1)}} A_{i,c}.
\]
The confidence margin was defined as
\[
m_i = b_i - b_i^{(2)}.
\]
A query image was considered a confident known-identity match only if
\[
b_i \geq \tau_{\mathrm{attach}}
\quad \text{and} \quad
m_i \geq \tau_{\mathrm{margin}}.
\]
The first condition ensured that the best known identity had sufficiently high similarity, while the second condition ensured that the assignment was not ambiguous with respect to the next-best identity.

\subsubsection{Cluster-Level Known-Identity Attachment}

Known-identity decisions were applied at the cluster level rather than directly at the image level. After Chinese Whispers produced query clusters, each confident query image cast a weighted vote for its best known identity. For a cluster $z$ and known identity $c$, the accumulated vote was
\[
V_{z,c}
=
\sum_{i:\, z_i=z}
\mathbb{1}
\left[
b_i \geq \tau_{\mathrm{attach}}
\land
m_i \geq \tau_{\mathrm{margin}}
\land
c_i^{(1)}=c
\right]
b_i,
\]
where $z_i$ was the cluster assigned to query image $q_i$. If at least one known identity received votes in cluster $z$, the cluster was assigned to the identity with the highest accumulated vote:
\[
\hat{c}_z = \arg\max_c V_{z,c}.
\]
All images in that cluster were then labeled as the selected known identity.

If a cluster received no confident known-identity votes, it was treated as a newly discovered individual. The final label for query image $q_i$ was therefore
\[
\hat{y}_i =
\begin{cases}
\texttt{cluster\_\{dataset\}\_\{}\hat{c}_{z_i}\texttt{\}},
&
\text{if cluster } z_i \text{ was attached to a known identity},\\
\texttt{cluster\_\{dataset\}\_new\_\{}z_i\texttt{\}},
&
\text{otherwise}.
\end{cases}
\]
For species without a labeled database, the query--database matrix was unavailable and all clusters were treated as newly discovered identities.

This cluster-level voting strategy made the final prediction more stable than direct image-level attachment. A single uncertain query image could not attach to a known identity unless it passed both the score and margin thresholds, and when multiple confident images occurred in the same cluster, their evidence was accumulated before assigning the final known label.

\subsection{MiewID Fine-Tuning}
Although the out-of-the-box pretrained MiewID model already produced strong results, we further adapted it to the AnimalCLEF26 setting through species-aware fine-tuning. In this work, we explored two fine-tuning strategies. The first strategy fine-tuned MiewID with a dynamic ArcFace objective to improve embedding discrimination, while the second strategy explored a fine-tuned variant based on SphereFace2 combined with focal loss.

\subsubsection{Dynamic ArcFace Fine-Tuning}

We fine-tuned the pretrained \texttt{MiewID-msv3} model to improve the discriminative quality of the global descriptor used in WildFusion. The goal of this stage was not closed-set identity classification, but learning a more compact and better-separated embedding space that remained useful for open-set identity discovery \cite{animalclef2026, vcermak2024wildlifedatasets}.

To approximate the challenge setting, we constructed an open-set species-wise split in which a subset of validation identities was excluded from fine-tuning. The labeled data used for this stage contained 13{,}074 images from 1{,}102 identities across lynx, salamander, and sea turtle datasets, while Texas horned lizard contributed no labeled identities because it was available only in the challenge test set. Table~\ref{tab:dynamicarcface_open_set_split} summarizes the resulting train/validation partition.

\begin{table}[!htbp]
\centering
\caption{Species-wise open-set split used for Dynamic ArcFace fine-tuning. The Train set (FT) contained individuals used for fine-tuning, while the Validation set was partitioned into ``Known'' and ``Unknown'' individuals to evaluate discovery performance.}
\label{tab:dynamicarcface_open_set_split}
\scriptsize
\setlength{\tabcolsep}{3.5pt}
\begin{tabular}{lrrccrrccrrrr}
\toprule
& \multicolumn{2}{c}{\textbf{Original Data}} & & \multicolumn{2}{c}{\textbf{Train Set (FT)}} & & \multicolumn{4}{c}{\textbf{Validation Set}} \\
\cmidrule{2-3} \cmidrule{5-6} \cmidrule{8-11}
\textbf{Species} & \textbf{imgs.} & \textbf{IDs} & & \textbf{imgs.} & \textbf{IDs} & & \textbf{Known IDs} & \textbf{Known imgs.} & \textbf{Unk. IDs} & \textbf{Unk. imgs.} \\
\midrule
LynxID2025        & 2957  & 77   & & 2519 & 75   & & 69  & 339  & 2   & 99  \\
SalamanderID2025  & 1388  & 587  & & 1087 & 558  & & 208 & 233  & 29  & 68  \\
SeaTurtleID2022   & 8729  & 438  & & 6265 & 419  & & 263 & 2136 & 19  & 328 \\
\midrule
Total             & 13074 & 1102 & & 9871 & 1052 & & 540 & 2708 & 50  & 495 \\
\bottomrule
\end{tabular}
\end{table}

We initialized the model from the pretrained \texttt{MiewID-msv3} checkpoint and fine-tuned it with a Dynamic ArcFace sub-center head using a two-phase schedule \cite{deng2019arcface, jiao2021dyn, tan2021efficientnetv2}. In the first phase, the backbone was frozen and only the classification head was trained as a short warm-up step. In the second phase, the full network was unfrozen and optimized end-to-end. In our configuration, the head used $k=3$ sub-centers per class, embedding dimension 2048, scale $s=64.0$, angular margin $m=0.6$, and threshold $\theta_0=0.785$. After fine-tuning, the classification head was discarded and only the normalized backbone embeddings were retained for similarity computation in the WildFusion pipeline.

Qualitatively, the resulting embeddings showed tighter within-identity clusters and clearer separation between neighboring identities than the pretrained model, which was consistent with the objective of improving downstream clustering and known-identity attachment in the open-set setting. This effect is shown in panel~(b) of Figure~\ref{fig:embedding_grid}, using a UMAP projection generated with HyperView software~\cite{mcinnes2018umap, hyperview2026}.

\subsubsection{SphereFace2-Focal Fine-Tuning}

We initialized the model from the pretrained \texttt{MiewID-msv3} checkpoint
\cite{otarashvili2024multispecies}
and adapted it as a feature extractor with an identity-discriminative training head.
All input images were resized to $440 \times 440$ pixels and normalized using ImageNet
statistics. For the salamander subset, we used segmented images composited on a
white background, motivated by the observation that background clutter can bias animal
re-identification models away from individual-specific visual traits
\cite{animalclef2026}.

To evaluate the fine-tuned MiewID model under an open-set setting, we constructed
a species-wise split in which known validation identities retained training images,
while unknown validation identities were completely removed from the fine-tuning set.
This allowed the model to be evaluated on both known-individual retrieval and
unseen-individual rejection, while preserving most of the available labeled data
for fine-tuning \cite{animalclef2026}. The resulting partition is reported in
Table~\ref{tab:miewid_open_set_split}.

\begin{table}[!htbp]
\centering
\caption{Species-wise open-set split used for SphereFace2-Focal fine-tuning. The Train set (FT) contained individuals used for fine-tuning, while the Validation set was partitioned into ``Known'' and ``Unknown'' individuals to evaluate discovery performance.}
\label{tab:miewid_open_set_split}
\scriptsize
\setlength{\tabcolsep}{3.5pt}
\begin{tabular}{lrrccrrccrrrr}
\toprule
& \multicolumn{2}{c}{\textbf{Original Data}} & & \multicolumn{2}{c}{\textbf{Train Set (FT)}} & & \multicolumn{4}{c}{\textbf{Validation Set}} \\
\cmidrule{2-3} \cmidrule{5-6} \cmidrule{8-11}
\textbf{Species} & \textbf{imgs.} & \textbf{IDs} & & \textbf{imgs.} & \textbf{IDs} & & \textbf{Known IDs} & \textbf{Known imgs.} & \textbf{Unk. IDs} & \textbf{Unk. imgs.} \\
\midrule
LynxID2025        & 2957  & 77   & & 2816 & 55   & & 19  & 38   & 22  & 103 \\
SalamanderID2025  & 1388  & 587  & & 1231 & 538  & & 59  & 59   & 49  & 98  \\
SeaTurtleID2022   & 8729  & 438  & & 8467 & 388  & & 50  & 100  & 50  & 162 \\
\midrule
Total             & 13074 & 1102 & & 12514 & 981 & & 128 & 197  & 121 & 363 \\
\bottomrule
\end{tabular}
\end{table}

\paragraph{SphereFace2-Focal Objective}

The fine-tuning head was based on the SphereFace2 binary-classification formulation
\cite{wen2022sphereface2binaryclassificationneed}. Instead of training the model as a
standard closed-set classifier, SphereFace2 treats each identity as an independent
one-vs-all decision. For each training image, the correct identity is treated as the
positive class, while all other identities are treated as negative classes. This design
is better aligned with open-set re-identification, because the final task depends on
pairwise similarity and unknown-individual discovery rather than choosing only from
a fixed list of training identities.

Following SphereFace2, the model learned normalized embeddings and normalized
identity prototypes on the hypersphere. The head used a similarity adjustment step to
make positive and negative similarities more separable, together with a two-sided
margin that made same-identity pairs satisfy a stricter similarity requirement while
pushing different-identity pairs farther from the decision boundary. A shared bias term
was used as a universal threshold across identities, making the binary decisions
more flexible and stable during training.

Because each image has one positive identity but many negative identities, we used
positive-negative reweighting to avoid the loss being dominated by negative classes.
We also combined the SphereFace2 objective with focal weighting
\cite{lin2017focal}. This reduced the influence of easy decisions and encouraged the
model to focus more on difficult cases, such as visually similar individuals from the
same species.

\begin{figure}[!htbp]
\centering
\begin{tabular}{@{}ccc@{}}
\includegraphics[width=0.31\linewidth,height=0.22\linewidth,keepaspectratio]{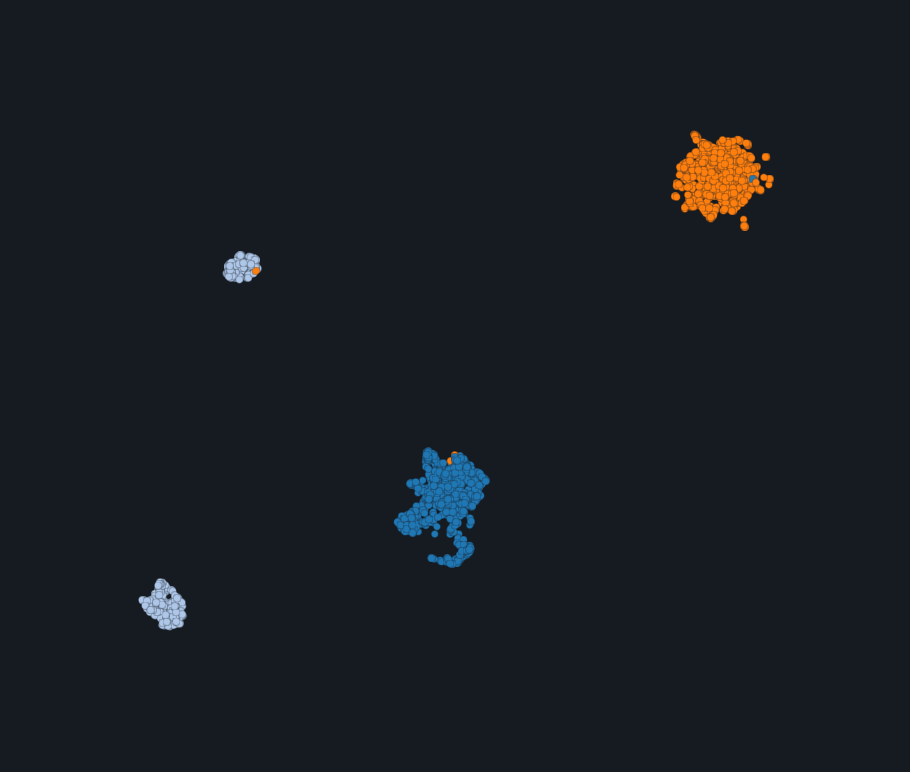} &
\includegraphics[width=0.31\linewidth,height=0.22\linewidth,keepaspectratio]{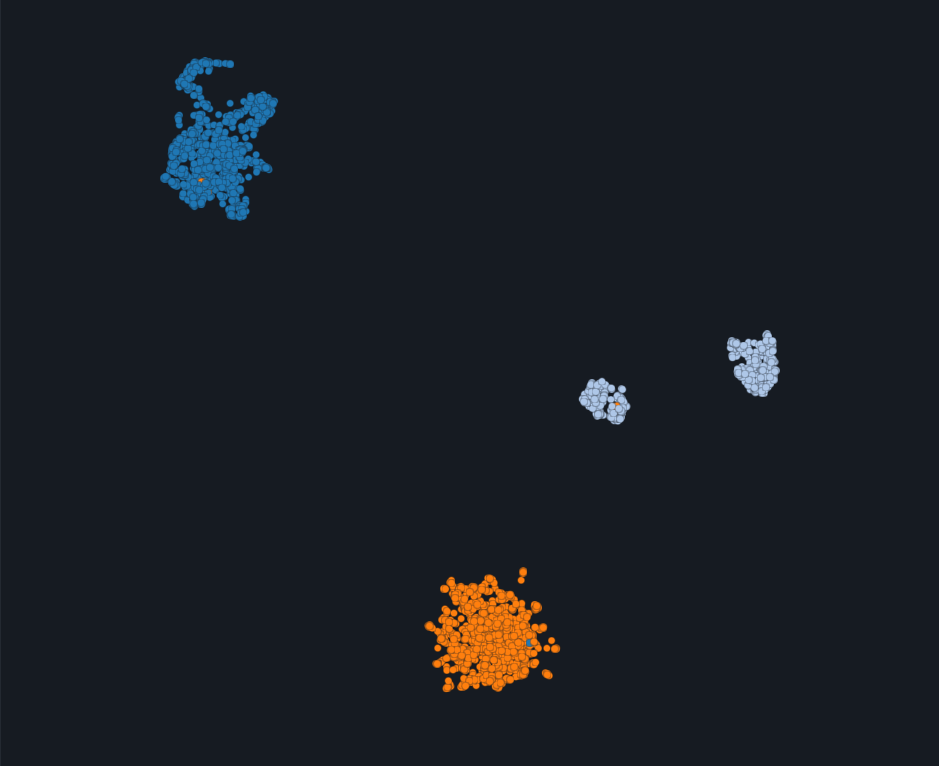} &
\includegraphics[width=0.31\linewidth,height=0.22\linewidth,keepaspectratio]{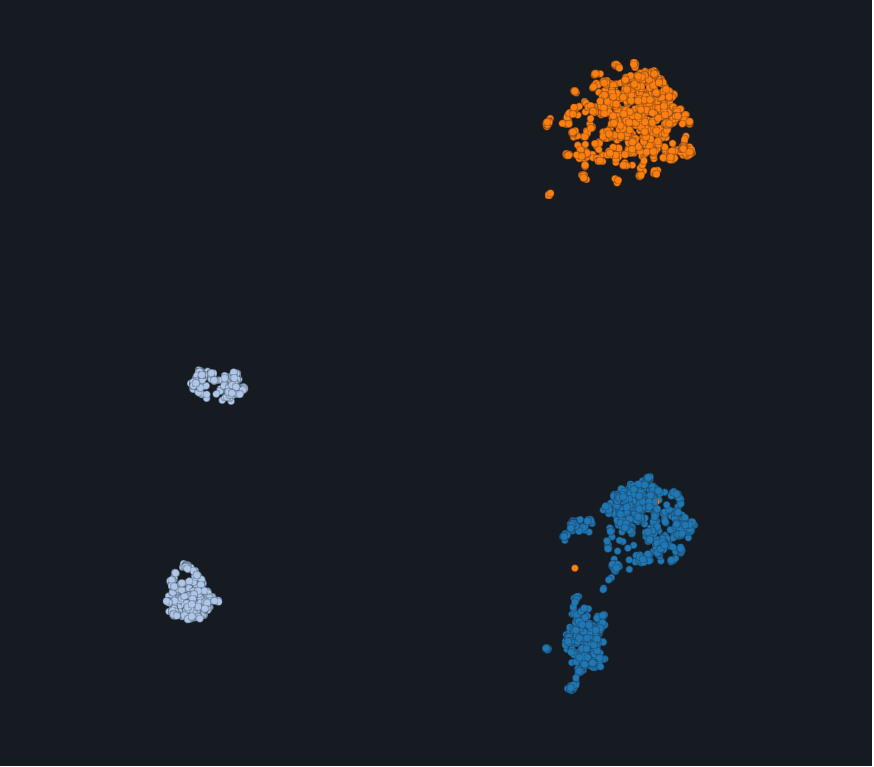} \\
\scriptsize (a) Original MiewID &
\scriptsize (b) Dynamic ArcFace &
\scriptsize (c) SphereFace2-Focal
\end{tabular}
\caption{Embedding-space visualizations for the Original MiewID model, Dynamic ArcFace fine-tuning, and SphereFace2-Focal fine-tuning. Colors denote datasets: blue is LynxID2025, orange is SeaTurtleID2022, and gray is SalamanderID2025.}
\label{fig:embedding_grid}
\end{figure}

In our experiments, we used a SphereFace2-Focal head with margin 0.3, scale 55,
similarity-adjustment strength 3.0, positive-negative balance 0.99, global loss
weight 20, and focal focusing parameter 1.0. After fine-tuning, the classification
head was discarded and only the normalized backbone embeddings were used. These
embeddings were then integrated into the WildFusion-based similarity pipeline as the
global descriptor, where improved embedding compactness could benefit both
known-identity attachment and unknown-individual clustering. Panel~(c) of Figure~\ref{fig:embedding_grid}
shows a similar tightening of the embedding space after fine-tuning, using a UMAP
projection generated with HyperView software~\cite{mcinnes2018umap, hyperview2026}.

\FloatBarrier

\subsection{Hyperparameter Optimization with Optuna}

We used Optuna~\cite{akiba2019optuna} to select the species-specific decision parameters of the similarity-to-clustering pipeline. Validation splits followed the open-set competition setting: identities were divided into known and unknown sets, and known identities were further split into reference and query images (70\% / 30\%).

To keep the search efficient, pairwise similarity matrices were precomputed and reused across trials. Each trial selected the known-identity scoring parameters $r$, $\tau_{\mathrm{attach}}$, and $\tau_{\mathrm{margin}}$; the re-ranking setting and parameters $(k_1,k_2,\lambda)$; and the Chinese Whispers graph parameters $(\tau_{\mathrm{CW}}, k_{\mathrm{NN}}, k_{\mathrm{core}}, T)$.

Each trial then followed the same inference pipeline described above. Query--database similarities were aggregated with top-$r$ pooling for known-identity scoring. The selected query--query matrix $S^{\mathrm{CW}}$ was built from either the re-ranked similarity matrix or the original query--query matrix, depending on the species-specific configuration, and Chinese Whispers was applied to the resulting sparse weighted graph.

After clustering, known-identity attachment was performed at the cluster level. Confident query images cast weighted votes when $b_i \geq \tau_{\mathrm{attach}}$ and $m_i \geq \tau_{\mathrm{margin}}$; clusters without confident votes were treated as newly discovered individuals. The validation objective was
\[
0.6 \times \mathrm{ARI} + 0.4 \times \mathrm{V\text{-}measure},
\]
Here, ARI was the dominant term because it matched the competition metric. We included V-measure only as a secondary regularizing term to discourage degenerate clusterings, such as merging multiple identities into one large cluster, while all official leaderboard scores were computed using ARI alone \cite{rosenberg2007vmeasure}. The best configuration was used in the final submission pipeline.

\subsection{Ensembling}
The pretrained MiewID, Dynamic ArcFace, and SphereFace2-Focal variants often produced complementary errors, so the final system combines their predicted labels by majority voting. For salamanders, most variants used segmented images composited on a white background, whereas the Dynamic ArcFace variant used a black background. The ensemble therefore benefits from both descriptor diversity and modest preprocessing variation.

\section{Evaluation Metric}

AnimalCLEF26 requires methods to both assign query images to known individuals and discover previously unseen identities. Consequently, submissions are evaluated using the \textbf{Adjusted Rand Index (ARI)}, which measures the similarity between the predicted clustering and the ground-truth clustering while correcting for agreement expected by chance \cite{animalclef2026,hubert1985comparing}.

Unlike the Rand Index (RI), which measures the proportion of image pairs on which the predicted and ground-truth clusterings agree, ARI normalizes this agreement so that random clusterings receive scores close to zero, while identical clusterings receive a score of one \cite{hubert1985comparing}.

The ARI is defined as
\begin{equation}
ARI=\frac{RI-E[RI]}{\max(RI)-E[RI]},
\end{equation}
where $E[RI]$ denotes the expected Rand Index under a random assignment model \cite{hubert1985comparing}.

Because ARI penalizes both over-segmentation (splitting one individual into multiple clusters) and under-segmentation (merging different individuals into the same cluster), it is well suited to the open-set animal re-identification task addressed in AnimalCLEF26.

\section{Results}
The main pipeline segments each specimen, applies additional species-specific preprocessing for lynx, sea turtle, and salamander, while using segmentation-only images for Texas horned lizard, and computes calibrated WildFusion similarities from a MiewID global descriptor plus ALIKED + LightGlue and DISK + LightGlue local branches. These similarities drive Chinese Whispers clustering, cluster-level known-identity attachment when labeled references exist, and the final ensemble across three MiewID variants. Table~\ref{tab:results} reports the official AnimalCLEF26 public and private leaderboard results using Adjusted Rand Index (ARI). We compare the competition baseline, which uses agglomerative clustering, individual pipeline variants, and our final ensemble submission.

\begin{table}[!htbp]
\centering
\caption{Comparison of ARI scores for the baseline, cumulative ablations, individual pipeline variants, and the final ensemble on the AnimalCLEF26 public and private leaderboards.}
\label{tab:results}
\small
\begin{tabular}{lcc}
\toprule
\textbf{Method} & \textbf{Public ARI} & \textbf{Private ARI} \\
\midrule
Competition baseline (WildFusion) & 0.20342 & 0.21221 \\
Baseline using only MiewID as global feature extractor & 0.60378 & 0.54901 \\
\quad + SAM segmentation  & 0.63996 & 0.57880 \\
\quad + Species-specific preprocessing & 0.64784 & 0.57611 \\
Main pipeline with Chinese Whispers clustering (out-of-the-box) & 0.71919 & 0.66695 \\
Dynamic ArcFace fine-tuned MiewID pipeline & 0.70508 & 0.63749 \\
SphereFace2-Focal fine-tuned MiewID pipeline & 0.70584 & 0.65736 \\
Final ensemble & \textbf{0.72124} & \textbf{0.70393} \\
\bottomrule
\end{tabular}
\end{table}

\begin{table}[!htbp]
\centering
\caption{Species-specific Optuna-selected hyperparameters for the three MiewID-based pipeline variants. The parameters control known-identity attachment, Chinese Whispers graph clustering, and k-reciprocal re-ranking.}
\label{tab:optuna_hyperparameters}
\scriptsize
\setlength{\tabcolsep}{2.5pt}
\begin{adjustbox}{max width=\linewidth}
\begin{tabular}{llccccccccc}
\toprule
\textbf{Pipeline variant} & \textbf{Dataset} 
& \textbf{top-$r$} 
& $\boldsymbol{\tau_{\mathrm{attach}}}$ 
& $\boldsymbol{\tau_{\mathrm{margin}}}$ 
& $\boldsymbol{\tau_{\mathrm{CW}}}$ 
& $\boldsymbol{k_{\mathrm{NN}}}$ 
& $\boldsymbol{T}$ 
& $\boldsymbol{k_{\mathrm{core}}}$ 
& \textbf{Re-rank} 
& $\boldsymbol{(k_1,k_2,\lambda)}$ \\
\midrule

\multirow{4}{*}{Original MiewID}
& LynxID2025        & 1 & 0.49 & 0.26 & 0.27 & 47 & 11 & 1 & Yes & (11, 3, 0.835) \\
& SalamanderID2025  & 2 & 0.25 & 0.19 & 0.50 & 25 & 12 & 1 & Yes & (5, 4, 0.758) \\
& SeaTurtleID2022   & 4 & 0.79 & 0.19 & 0.25 & 47 & 15 & 1 & Yes & (39, 3, 0.449) \\
& TexasHornedLizards& \textemdash & \textemdash & \textemdash & 0.49 & 15 & 20 & 1 & No & \textemdash \\

\midrule

\multirow{4}{*}{SphereFace2-Focal MiewID}
& LynxID2025        & 1 & 0.28 & 0.42 & 0.29 & 42 & 10 & 1 & Yes & (10, 1, 0.808) \\
& SalamanderID2025  & 2 & 0.35 & 0.23 & 0.40 & 60 & 20 & 5 & Yes & (7, 4, 0.718) \\
& SeaTurtleID2022   & 5 & 0.82 & 0.45 & 0.23 & 47 & 25 & 1 & Yes & (20, 5, 0.852) \\
& TexasHornedLizards& \textemdash & \textemdash & \textemdash & 0.49 & 15 & 20 & 1 & No & \textemdash \\

\midrule

\multirow{4}{*}{Dynamic ArcFace MiewID}
& LynxID2025        & 5 & 0.29 & 0.17 & 0.24 & 57 & 17 & 3 & Yes & (5, 4, 0.810) \\
& SalamanderID2025  & 3 & 0.19 & 0.24 & 0.43 & 55 & 29 & 1 & Yes & (5, 9, 0.662) \\
& SeaTurtleID2022   & 5 & 0.51 & 0.80 & 0.21 & 49 & 7  & 1 & Yes & (9, 1, 0.824) \\
& TexasHornedLizards& \textemdash & \textemdash & \textemdash & 0.49 & 15 & 20 & 1 & No & \textemdash \\

\bottomrule
\end{tabular}
\end{adjustbox}
\end{table}

Table~\ref{tab:optuna_hyperparameters} reports the species-specific hyperparameters selected by Optuna for each MiewID-based pipeline variant. The first three species use full query--query clustering followed by cluster-level known-identity attachment, while Texas horned lizard has no labeled database and therefore uses only the clustering-related parameters.

\subsection{Additional Submitted Configuration: Preprocessing Before Calibration}

We also evaluated an additional competition-time configuration to study the effect of applying the additional species-specific preprocessing steps before WildFusion calibration. Unlike the final ensemble, this configuration used only the original pretrained \texttt{MiewID-msv3} model as the global descriptor, while keeping the same ALIKED + LightGlue and DISK + LightGlue local branches.

The main difference was the order of preprocessing and calibration. In this configuration, calibration was fitted on the same preprocessed image distribution used during final inference:
\[
\text{segmentation}
\rightarrow
\text{species-specific preprocessing}
\rightarrow
\text{WildFusion calibration}
\rightarrow
\text{WildFusion inference}.
\]

\begin{table}[!htbp]
\centering
\caption{Comparison between the final selected ensemble submission and the additional submitted configuration on the official AnimalCLEF26 leaderboards.}
\label{tab:additional_config_comparison}
\begin{tabular}{lcc}
\toprule
\textbf{Configuration} & \textbf{Public ARI} & \textbf{Private ARI} \\
\midrule
Final ensemble submission & \textbf{0.72124} & 0.70393 \\
Preprocessing before calibration & 0.71439 & \textbf{0.71087} \\
\bottomrule
\end{tabular}
\end{table}

As shown in Table~\ref{tab:additional_config_comparison}, the final ensemble achieved a higher public score, while the preprocessing-before-calibration configuration achieved a higher private score. This suggests that fitting WildFusion calibration on the final image distribution, including species-specific preprocessing where used, may improve hidden-split generalization. It is also computationally simpler because it avoids fine-tuned descriptor variants and multi-model ensembling.

\section{Discussion}
The results show that most of the performance gain comes from the overall similarity-to-clustering design rather than from fine-tuning alone. Even the training-free pipeline strongly outperforms the competition baseline, indicating that specimen segmentation, targeted species-specific preprocessing where beneficial, and calibrated fusion of global and local cues already provide a strong solution for discovery-oriented re-identification.

The fine-tuned MiewID variants remain useful because they add complementary information, even though they do not outperform the training-free model individually. This is reflected in the ensemble, which yielded our best selected submission on the public leaderboard, while the preprocessing-before-calibration configuration achieved the strongest private leaderboard score. Together, these findings suggest that representation diversity improves robustness and that calibration strategy remains a key factor for generalization.

\section{Limitations}

Despite its strong performance, the proposed system has three main limitations. First, our pipeline uses Chinese Whispers to convert the query--query similarity graph into identity clusters. Although efficient and well suited to open-set discovery, Chinese Whispers updates node labels in randomized order, so repeated runs can produce slightly different cluster assignments, especially for weakly connected or visually ambiguous samples. This limits reproducibility relative to deterministic graph-clustering methods \cite{biemann2006chinese}.

Second, the complete system is computationally expensive. Local matching and pairwise similarity computation require comparing many image pairs, and the final ensemble repeats the pipeline for three MiewID-based descriptor variants, increasing both inference time and memory usage.

Third, the calibration stage was fitted using only Lynx images rather than a more diverse multi-species calibration set. This may limit how well the calibrated similarities transfer across species with different visual statistics and imaging conditions.

These limitations suggest several directions for future work, including more deterministic graph clustering, lighter single-model alternatives that reduce the cost of pairwise matching and ensembling, and broader calibration strategies that better cover species diversity. The latter is especially promising because a simpler off-the-shelf configuration remained competitive on the private leaderboard.

\section{Conclusion}
We presented a discovery-oriented animal re-identification pipeline for AnimalCLEF26 that combines specimen segmentation, targeted species-specific preprocessing where beneficial, WildFusion-based similarity estimation, graph clustering, and known-identity attachment. The training-free pipeline already performs strongly, while combining training-free and fine-tuned MiewID variants yields our best public leaderboard submission and highlights the value of complementary global-local cues; however, a simpler preprocessing-before-calibration configuration achieved our strongest private leaderboard score. Future work will focus on more deterministic clustering, more efficient similarity computation, and improved calibration for stronger generalization.

\begin{acknowledgments}
We thank the LifeCLEF organizers for making the AnimalCLEF 2026 datasets available and for organizing the competition. We are especially grateful to Eslam Yacoub, PhD, Assistant Professor of English Curricula and Instruction at Alamein International University, for carefully reviewing the manuscript and for his guidance throughout the preparation of this working note. We also thank Antonio Rueda-Toicen from Hasso Plattner Institute for his support and thoughtful review of the paper, and Bahey Tharwat, PhD student at the University of Freiburg, for reviewing the paper and helping with figure preparation.
\end{acknowledgments}

\section*{Declaration on Generative AI}
During the preparation of this work, the authors used ChatGPT-5.5 and Prism GPT-5.2 for grammar and spelling checks. After using these tools, the authors reviewed and edited the content as needed and assumed full responsibility for the final version of the work.
\bibliography{sample-ceur}

\end{document}